\documentclass[12pt]{iopart}

\usepackage{graphicx}
\usepackage{amsfonts}
\usepackage{color}
\usepackage{subcaption}

\newcommand{\muv}{\mbox {\boldmath $\mu$}}

\begin{document}

\title[Data-driven-based roughness descriptors discovery and consolidation modeling]{Data-driven discovery of roughness descriptors for surface characterization and intimate contact modeling of unidirectional composite tapes}

\author{Sebastian Rodriguez$^1$, Mikhael Tannous$^1$, Jad Mounayer$^1$, Camilo Cruz$^2$, Anais Barasinski$^2$ and Francisco Chinesta$^{1,2}$}

\address{$^1$ AI4Engineering Task Force @ PIMM Lab, Arts et Metiers Institute of Technology, Paris, France \\
$^2$ Univ Toulouse, IMT Mines Albi, INSA Toulouse, ISAE-SUPAERO, CNRS, ICA, Albi, France}
\ead{Anais.Barasinski@mines-albi.fr \ Francisco.Chinesta@ensam.eu}
\vspace{10pt}

\begin{abstract}
Unidirectional tapes surface roughness determines the evolution of the degree of intimate contact required for ensuring the thermoplastic molecular diffusion and the associated inter-tapes consolidation during manufacturing of composite structures. However, usual characterization of rough surfaces relies on statistical descriptors that even if they are able to represent the surface topology, they are not necessarily connected with the physics occurring at the interface during inter-tape consolidation. Thus, a key research question could be formulated as follows: {\em Which roughness descriptors simultaneously enable tape classification—crucial for process control—and consolidation modeling via the inference of the evolution of the degree of intimate contact, itself governed by the process parameters?}. For providing a valuable response, we propose a novel strategy based on the use of Rank Reduction Autoencoders (RRAEs), autoencoders with a linear latent vector space enforced by applying a truncated Singular Value Decomposition (SVD) to the latent matrix during the encoder-decoder training. In this work, we extract useful roughness descriptors by enforcing the latent SVD modes to (i) accurately represent the roughness after decoding, and (ii) allow the extraction of existing \emph{a priori} knowledge such as classification or modelling properties.
\end{abstract}

%
\vspace{2pc}

\section{Introduction}

Many composite forming processes for manufacturing structural parts are based on the consolidation of pre-impregnated preforms, e.g., sheets, tapes, .... To obtain mechanical cohesion between thermoplastic pre-impregnated preforms that are brought together during manufacturing, two specific physical conditions at the interface are needed: (a) the development of an effective intimate contact between surfaces that is a requirement for (b) the temperature-driven macro-molecular inter-diffusion (as described by De Gennes reptation theory \cite{deGennes}) within the process time window, while avoiding thermal degradation.
To reach this goal, both pressure and heat are usually applied during the manufacturing process. Determining the process window for reaching an optimal consolidation without degrading the material, is of critical importance for ensuring the highest productivity while certifying the quality of the resulting structure \cite{tannous2016thermo}. 

Tapes surface roughness determines the evolution of the degree of intimate contact, which is a sine qua non for the thermoplastic macro-molecular inter-diffusion between tapes. The extent of macro-molecular inter-diffusion dictates the mechanical properties of the polymer-polymer interface and therefore the mechanical integrity of the final composite structure. Typically, the roughness of a composite tape surface is characterized using classical statistical descriptors that even if they are able to represent the surface topology itself, there is not proof that those descriptors are the more suitable ones to describe the physical phenomena taking place during the intimate contact development between tapes.

Thus, the following research question appears pertinent:{\em Which roughness descriptors simultaneously enable tape classification—crucial for process control—and consolidation modeling via the inference of the evolution of the degree of intimate contact, itself governed by the process parameters?}.

Our response to this question follows a novel procedure based on the use of a powerful autoencoder with a linear latent vector space \cite{JAD,ben2026rank,idrissi2025new}, enforced by applying a truncated Singular Value Decomposition (SVD) at the latent space level during the encoder-decoder training. The resulting SVD modes are enforced to represent any existing {\em a priori} knowledge, for instance valuable statistical descriptors, complemented with the ones needed for inference purposes (classification or modeling), and finally for those needed to ensure the profiles representation after decoding.

\begin{figure}[h]
\centering
\begin{subfigure}{0.45\textwidth}
    \centering
    \includegraphics[width=\textwidth]{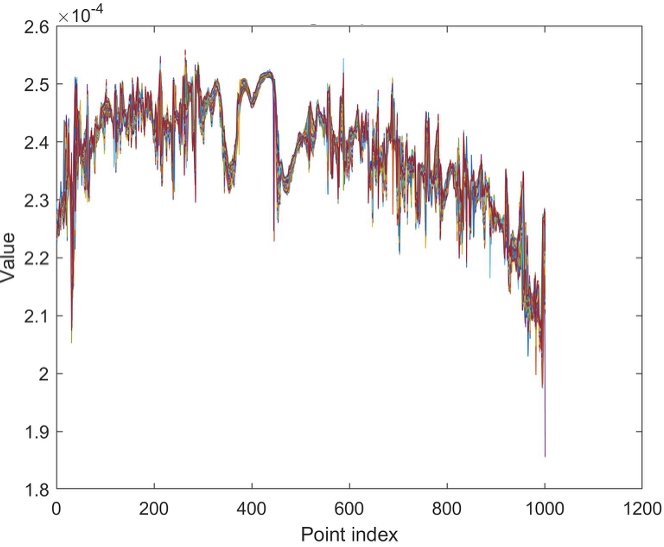}
    \caption{Macro-roughness profile.}
\end{subfigure}
\hfill
\begin{subfigure}{0.45\textwidth}
    \centering
    \includegraphics[width=\textwidth]{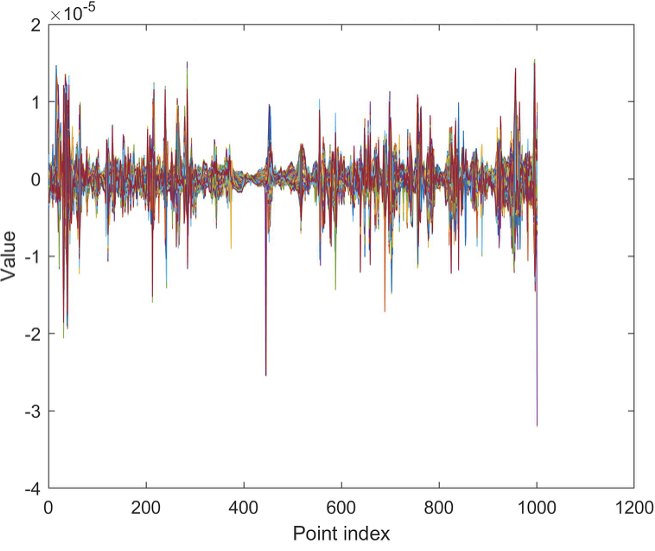}
    \caption{Associated micro-roughness after removing the dominant geometrical feature.}
\end{subfigure}
\caption{Illustration of macro- and micro-roughness measured on a unidirectional composite tape.}
\label{Fig:MacroMicro}
\end{figure}

To demonstrate the potential of the proposed technology, we use a database of hundreds of roughness profiles measured on 12 thermoplastic composite tapes produced by different providers. Conventional statistical descriptors were unable to reliably classify the surfaces or predict the evolution of intimate contact during consolidation. Topological Data Analysis (TDA) enabled classification in previous work \cite{runacher2023rough}, but its success mainly relied on the pronounced differences in macro-roughness between tapes from different suppliers. These large-scale geometrical features generate persistent topological signatures. As illustrated in Fig.~\ref{Fig:MacroMicro}, once the macro-scale component is removed, the resulting micro-roughness no longer exhibits clear persistent structures. Consequently, TDA loses its categorization capability when applied to the micro-roughness alone.

On the other hand, the data of intimate contact evolution are obtained in-silico by using a 2D simulation of the composite tape surface compaction. A cellular automaton computational method is used to simulate the compaction of the surface roughness profile at constant velocity. Each measured roughness profile is used as initial geometrical domain for a compaction simulation, which produces as output the time-resolved evolution of the degree of intimate contact.

The technology proposed in this paper operates with a single latent mode for classification when both macro- and micro-roughness scales are considered, and with four modes when focusing exclusively on the micro-roughness component. The micro-roughness is obtained by removing the macro-scale contribution from the measured surface profiles. As such, the proposed framework constitutes an unprecedented characterization and modeling tool of particular relevance for the community working on composite forming based on pre-impregnated preforms. Remarkably, the same number of latent modes—and only one additional mode when higher accuracy is required—also enables accurate prediction of the evolution of intimate contact, while simultaneously ensuring an almost perfect reconstruction of the roughness profile itself. This demonstrates the strong compactness and physical consistency of the reduced representation, capable of capturing geometry and consolidation kinetics within a unified low-dimensional space.

This article is organized into seven sections, including the present introduction. Section~\ref{sec:sim} presents in detail the numerical method to simulate the composite tape surface compaction, followed by the description of the database of composite tape roughness profiles in Section~\ref{sec:material}. Section~\ref{sec:class} details the rough surface classification methodology based on Rank Reduction Auto-Encoders (RRAE) and associated results. Section~\ref{sec:consolidation} introduces the intimate contact prediction using RRAE along with its associated results, before comparison with standard machine learning architectures in Section~\ref{sec:MLStand}. Finally, Section~\ref{sec:conclusion} provides conclusions and perspectives.

\section{Compaction simulation}\label{sec:sim}

In former works \cite{Leon2015,Leon2017,C2} we proposed simulating the consolidation of as-measured surfaces geometries instead of the sometimes crude reconstructions of them, (for example, based on the use of fractal features or the description of asperities by means of rectangular elements \cite{Yang2001,LEV}). A numerical strategy to geometrically model an as-measured surface roughness profile considers a Haar-based wavelet representation \cite{C2}, which results in a multi-scale sequence of rectangular patterns from the coarsest scale to the finest one. This geometrical model results in a high-fidelity representation of the as-measured surface. The smoother is the surface, the less levels in the wavelet-based description are required.
 
Given a tape surface geometrical representation, different simulation approaches can be envisaged to emulate the physical phenomena taking place during inter-tape consolidation. For example, in \cite{FRA} lubrication flow was considered to compress the rectangular elements (considering a 2D model of a surface profile) from the finest to the coarsest scales, while coupling with the thermal problem that determines the temperature field evolution, intimately coupled with the polymer viscosity. The space-separated representation of this problem enables the use of the so-called Proper Generalized Decomposition (PGD) \cite{C3,C4,C5,C6}, which allows to solve the 2D transient heat problem as a sequence of two 1D boundary-value problems defined in the two spatial coordinates respectively, and an ordinary differential equation for computing the temporal functions determining the separated representation solution in time. 

Considering unidirectional reinforcement in a pre-impregnated composite tape (pre-preg), one can distinguish two characteristic roughness directions that correspond to the physical dimensions of the tape surface. The roughness in the transversal direction exhibits smaller characteristic lengths in comparison with the longitudinal direction. This anisotropy in roughness can be explained by the presence of a unidirectional continuous fiber thread, where the roughness in the transversal direction is dictated by the mean fiber-fiber distance (close to the fiber diameter) and the roughness in the longitudinal direction follows the apparition of fiber crossing/overlapping defects. The characteristic length of those rigid fiber crossing/overlapping events should be at least 1-2 orders of magnitude larger that the fiber diameter. Based on the previous reasoning, we can assume that the tape surface compaction is driven mainly by the roughness in the transversal direction. The roughness along the longitudinal direction remains of second order with respect to the one in the transverse direction and could be neglected. This allows to reduce the dimensionality of the compaction problem by taking into account exclusively the roughness profile in the transversal direction of the composite tape. With this choice we implicitly neglect mass and heat transfer in the longitudinal direction.

\begin{figure}[!h]
\centering
\includegraphics[width=8 cm]{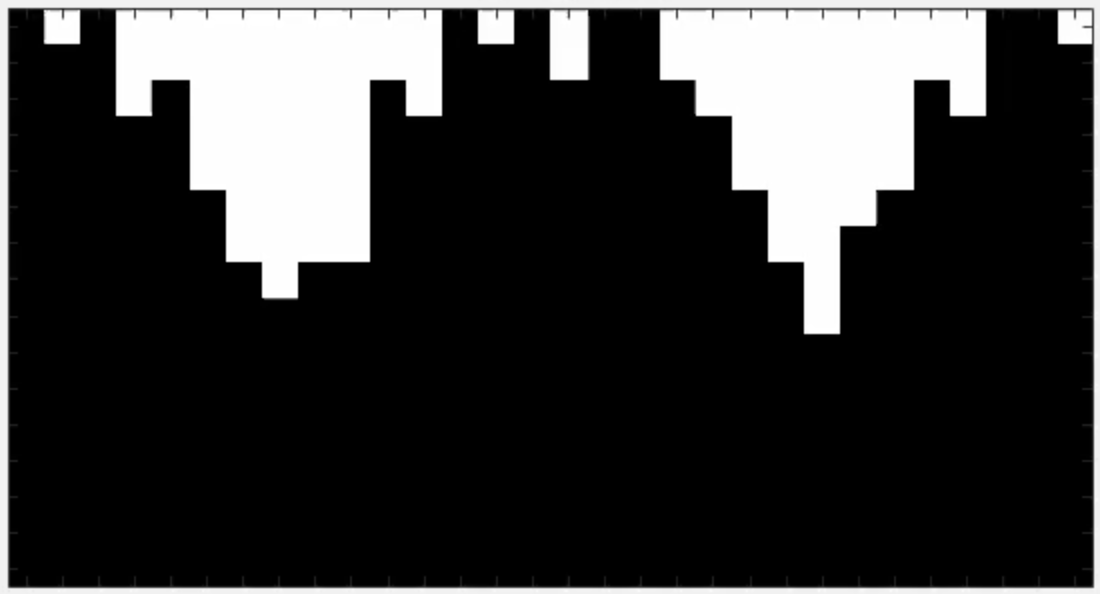}
\caption{Cellular automaton discretization of the 2D composite tape rough surface before compaction. Black cells represents composite tape while white represents air.}
\label{CA}
\end{figure}

To evaluate the evolution of degree of intimate contact, we simulate the isothermal compaction of a composite surface tape by an ideal-even rigid compression plate moving at constant velocity. We chose a cellular automaton computational method to solve this problem. This method requires a discretization of the space domain in a regular grid of cells. To represent the tape cross-section with its as-measured roughness, two cell states are considered: Composite tape material and air between tape surface and compression plate. A sketch of the cellular automaton discretization for this problem is given in Fig. \ref{CA}, where black cells represent the composite tape domain and the white cells represent air.

\begin{figure}
\centering
\includegraphics[width=6 cm]{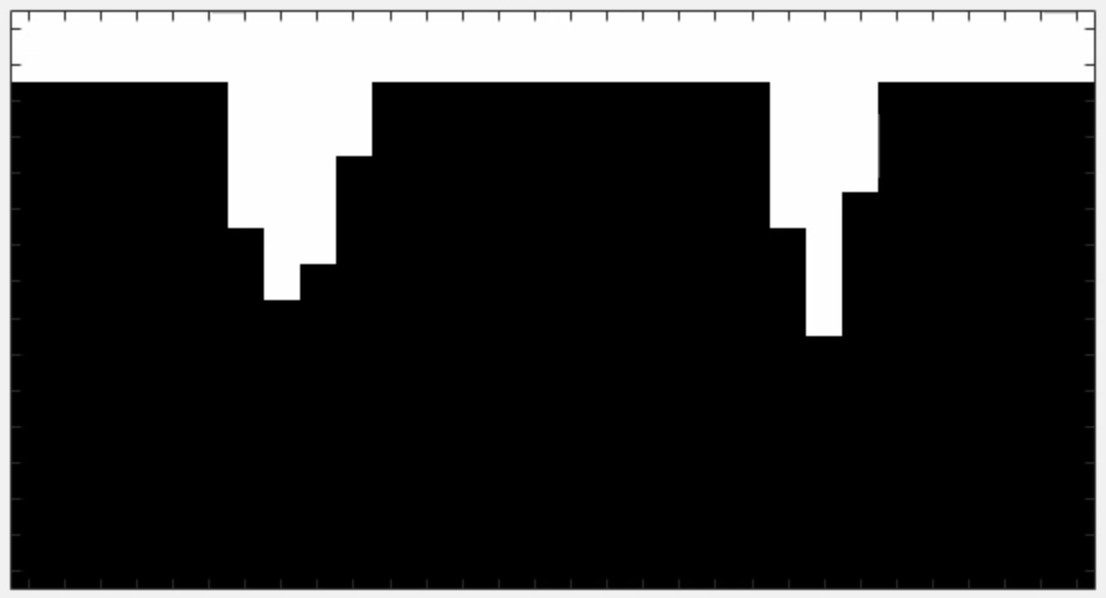}
\includegraphics[width=6 cm]{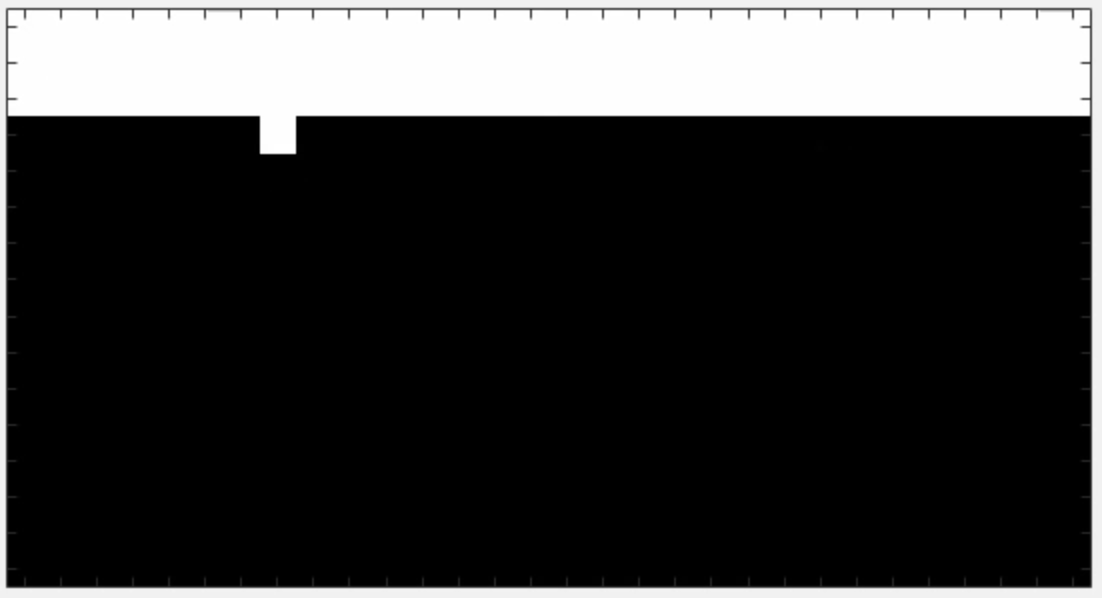}
\caption{Cellular automaton states of the 2D composite tape rough surface compaction: Intermediate step (left) and final (right) step.}
\label{CIF}
\end{figure}

During a compaction time step each composite cell in contact with the compression plate moves to the closest available air cell that has no contact with the compression plate. The underlying hypotheses in this numerical approach are the following: 1) moving cells are soften thermoplastic, 2) soften thermoplastic behaves as an incompressible transversally isotropic fluid, and 3) air is not compressed, but goes out of the domain (ideal venting). Figure \ref{CIF} depicts an intermediate compaction state as well as the final one for the initial rough surface sketched in Fig. \ref{CA}. 
The ratio of composite length in contact with the compression plate to the full tape width gives the instantaneous degree of intimate contact (DIC). Each cellular automaton simulation delivers therefore an estimation of the evolution of the DIC during compaction.

In terms of data, to each as-measured tape surface roughness profile $h(x)$, with $h(x)$ representing the surface height at coordinate $x$ along the tape transverse direction, corresponds an evolution of the degree of intimate contact $\mbox{DIC}(t)$. The latter constitutes an indicator of the capability of a given composite tape to reach high levels of inter-tape consolidation.

If one considers a number of profiles $h^j(x)$, $j=1, ..., P$, with $P$ the number of profiles, that at its turn could be expressed in a discrete manner from vector $\mathbf h^j$, whose $i$-component represents the surface height $h$ at location $x_i$, $i=1, ..., N$, assumed uniformly distributed along the profile length $L$ (tape width), with $x_i = (i-1) \frac{L}{N-1}$. Thus,  the $i$-component of vector $\mathbf h^j$, $\mathbf h^j_i$, $\forall j$, represents $h^j(x_i)$, i.e. $\mathbf h^j_i \equiv h^j(x_i)$, $i=1, ..., N$, $j=1, ..., P$.

Following the same rationale, by considering a time interval of length $T$, and $M$ time instants uniformly distributed in it, $t_k = (k-1) \frac{T}{M-1}$, $k=1, ..., M$, the continuous $\mbox{DIC}^j(t)$ related to the $j$-profile, can be expressed in a discrete manner from vector $\mathbf D^j$, whose $k$-component, $\mathbf D^j_k$, represents the DIC of the $j$-profile at time $t_k$.

In other words, the compaction simulation of the $P$ profiles $\mathbf h^j$, $j=1, ..., P$ results into the $P$ DIC evolution curves $\mathbf D^j$, $j=1, ..., P$, with $\mathbf h^j \in \mathbb R^N$ and $\mathbf D^j \in \mathbb R^M$.


\section{Database of roughness profiles}\label{sec:material}

\begin{figure}
\centering
\centerline{\includegraphics[width=15 cm]{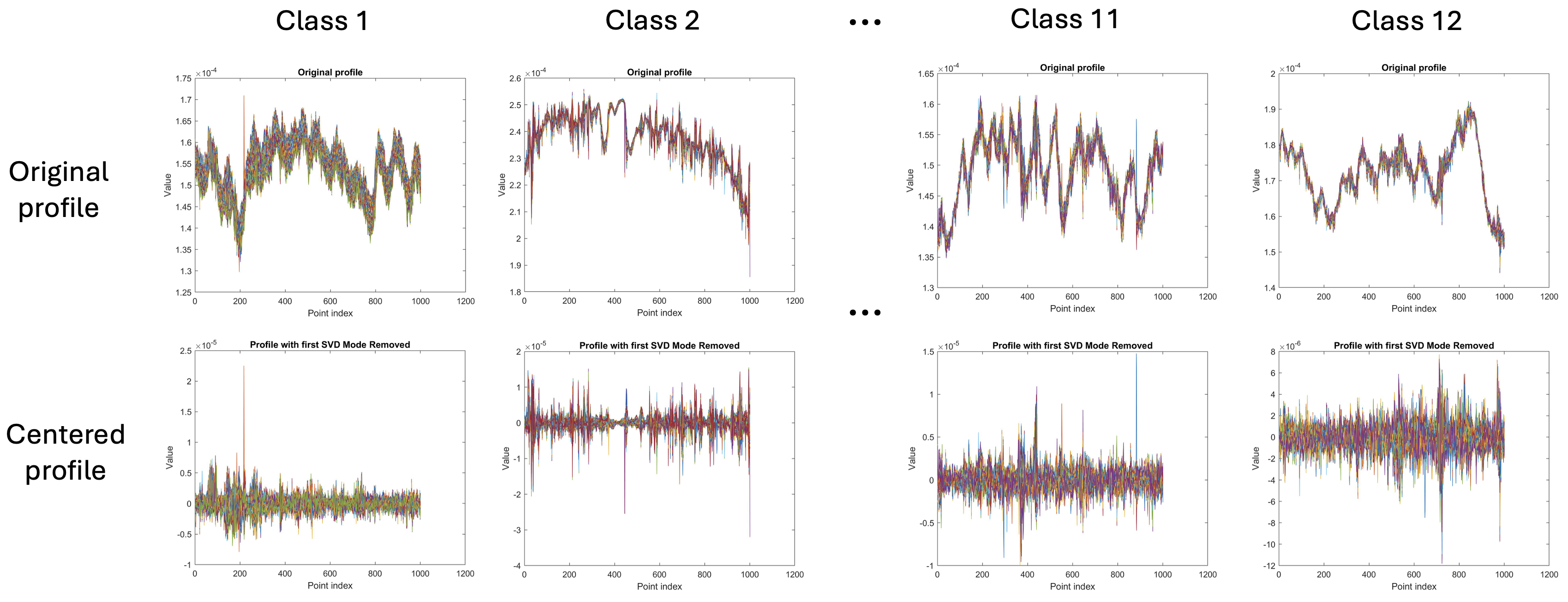}}
\caption{Different roughness profiles extracted from 12 tapes associated with different providers, here refereed as classes}
\label{Sur}
\end{figure}

As database for this work, we consider samples scanned with a 3D non contact profilometer, with a $3.5\, \mu$m resolution and where each sample is approximately 3 mm along the tape width. A set of 1011 surface profiles were extracted from 12 different thermoplastic composite unidirectional tapes coming from different suppliers. Each profile is represented by 1000 measured data points. Some of the as-measured roughness profiles in the transversal direction of the composite tape are shown in Fig.~\ref{Sur}. 

In this case, the use of classical roughness statistical descriptors was unable to classify or to predict the degree of intimate contact evolution during compaction. On the other hand, as discussed in \cite{FRA}, topological data analysis --TDA-- performed quite well for classifying the different composites with respect to their classes (providers).

TDA is based on extracting the topological structure of data \cite{TDA1,TDA2,TDA3,TDA4}. In \cite{FRA}, a discrete representation of the roughness profile $\mathbf{h}^j$ was considered, and for each vector, pairs of nearest local minima and maxima were identified. These pairs capture the persistent topological features of the signal and can be used directly, or through their derived persistence images, to characterize the surface roughness. In \cite{FRA}, persistence images were further employed for classification purposes, demonstrating excellent performance.

However, as Fig.~\ref{Sur} points out, the roughness involves two different scales, a macroscopic one, and the so-called micro-roughness, which results from removing the macroscopic roughness from the as-measured profile (bottom row in Fig.~\ref{Sur}). When trying to classify the micro-roughness, TDA fails due to the fact that the micro-roughness involves a non-persistent topology, which is basically noise from the topological point of view.

\begin{figure}
\centering
\centerline{\includegraphics[width=0.4\textwidth]{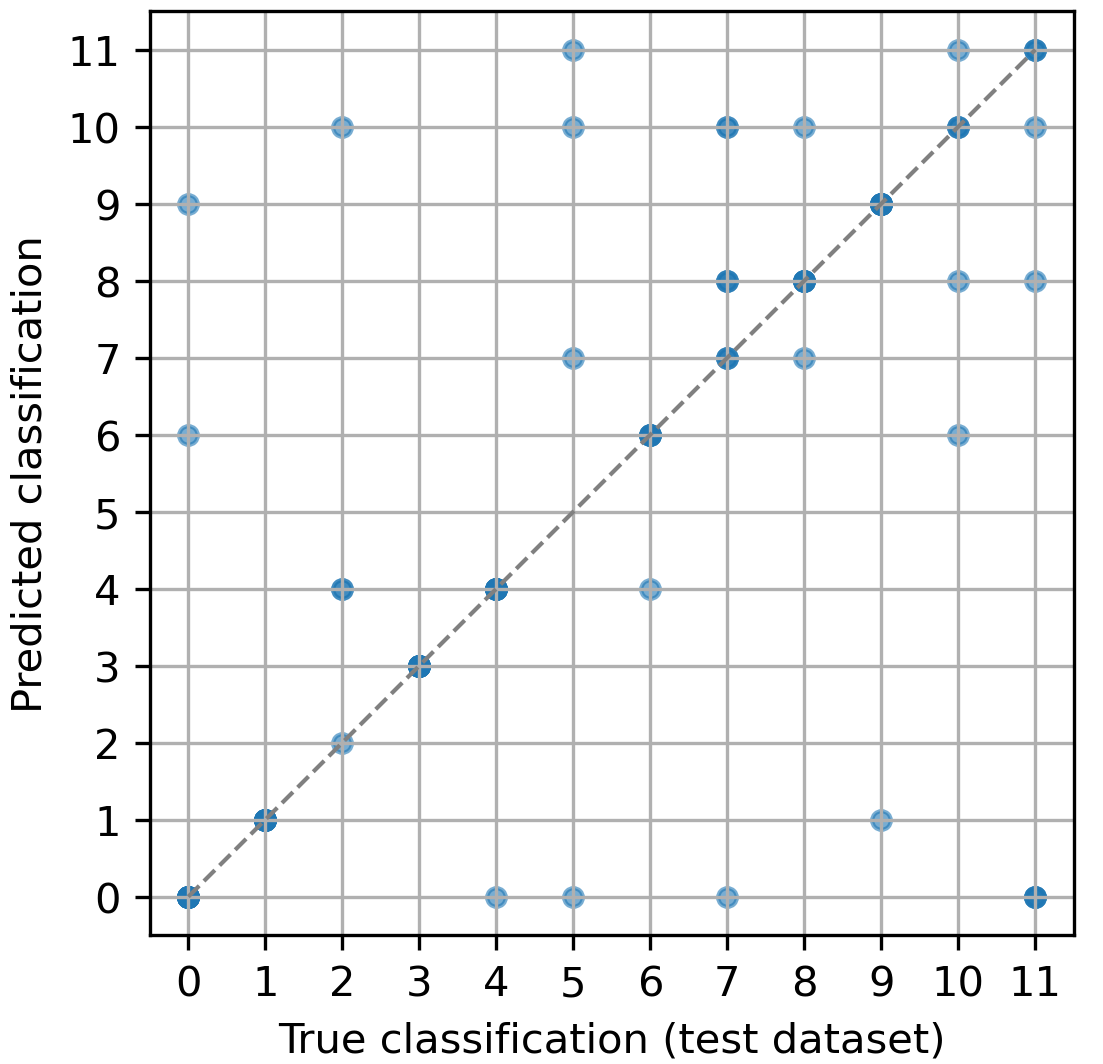}}
\caption{Classification accuracy obtained using Topological Data Analysis (TDA) based on persistence images.}
\label{fig:TDAClass}
\end{figure}

In order to illustrate the TDA's performance, we consider $1011$ roughness profiles, from which $911$ were used for training and the remaining $100$ for testing. Each roughness profile was first mapped to a phase-space representation using a Takens delay embedding. Persistent topology was then computed from the resulting trajectories using a Vietoris--Rips filtration, producing persistence diagrams describing the birth and death of topological features. These diagrams were subsequently vectorized using persistence images in order to obtain fixed-length feature vectors for classification. Alternative diagram summaries including persistence landscapes and persistence entropy were also evaluated.
When applied to the micro-roughness signals, the classification performance did not exceed $70\%$ accuracy on the test set (Fig.~\ref{fig:TDAClass}). This behavior can be explained by the fact that the micro-scale roughness is interpreted by TDA as topological structures with very low persistence. Consequently, most of the birth--death pairs concentrate near the diagonal of the persistence diagram, indicating topological features that behave essentially as noise and do not provide sufficient discriminative information for reliable classification. This observation suggests that the micro-roughness is dominated by non-persistent topological structures, which limits the effectiveness of standard TDA descriptors for this classification task.

In the following, we introduce a novel technology that succeeds in classifying at both, macro and micro, roughness scales.

\section{Rough surface classification}\label{sec:class}

\subsection{Data-reduction and Rank Reduction Autoencoder --RRAE}
Usual autoencoders are designed for reducing the data by removing linear and nonlinear correlations.

Thus, one can expect that the profiles $\mathbf h \in \mathbb R^N$ could be represented in a lower dimensional space $\mathbf y \in \mathbb R^n$, with $n \ll N$. 

For that purpose, it suffices looking for the mapping $\mathbf h \rightarrow \mathbf y(\mathbf h)$, that is in general nonlinear, and unknown. One could be tempted of representing it by a Neural Network to be trained from data, that is
$\mathbf y = \mathcal N^e_{\theta^e}(\mathbf h)$, where the superscript $\cdot^e$ refers to the encoding, and the index $\theta^e$ represent the set of trainable parameters related to the encoding neural network $\mathcal N^e$.

In order to guarantee that the reduction never removes critical information avoiding the reconstruction capabilities, a decoding mapping, performed by the so-called decoding neural network $\mathcal N^d_{\theta^d}$, with its associated trainable parameters $\theta^d$ is learnt, in such a way that the loss function is minimized:

\begin{equation}
\{ \theta^e, \theta^d\} = \mathop{\mathtt{argmin}}_{(\theta^{e*}, \theta^{d*})} \sum_{j=1}^P \frac{\| \mathcal N^d_{\theta^{d*}} \left ( \mathcal N^e_{\theta^{e*}} (\mathbf h^j) \right ) -\mathbf h^j \|_2}{\| \mathbf h^j \|_2}
\end{equation}
while trying to keep $n$ as small as possible. 

The training performs from the $P$ available data $\mathbf h^j$, $j=1, ..., P$. 

Here, the latent space, $\mathbf y= \mathcal N^e_{\theta^e} (\mathbf h)$, of reduced dimension $n$, $\mathbf y \in \mathbb R^n$ represents the data's intrinsic dimensionality. 

\begin{figure}
\centering
\includegraphics[width=0.75\linewidth]{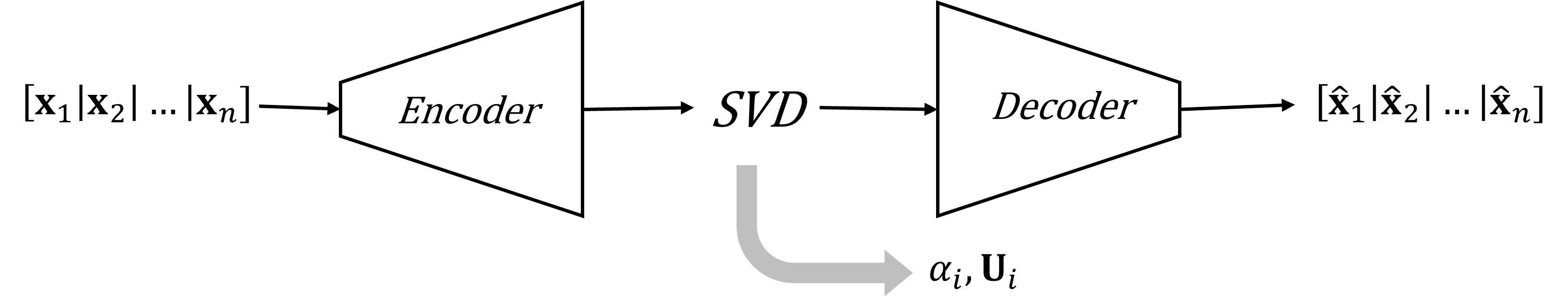} 
\caption{Rank Reduction Autoencoder. \label{fig:RRAE}}
\end{figure}

To improve expressibility, knowledge assimilation and interpolation (making possible data generation), a new autoencoding architecture was proposed in \cite{JAD}, called Rank Reduction Autoencoder (RRAE). This architecture consists of enforcing a linear vector structure into the latent space, from the use of a SVD intercalated in between encoder and decoder as depicted in Fig.~\ref{fig:RRAE}. An input vector $\bf{x}_j$ entering the RRAE will be reconstructed into $\hat{\bf{x}}_j$ while recovering at the SVD level, modes $\bf{U}_i$ ($i \in [1, k_{max}]$) associated with reduced descriptors $\alpha_i$. For more in depth information on the RRAE, the interested reader is referred to \cite{JAD}.

Thus, when using the data $\mathbf h^j$, one recovers at the latent space level a number of SVD modes $\mathbf U_1, ..., \mathbf U_{k_{max}}$, such that any latent representation of a roughness profile could be approximated in that basis:

\begin{equation}
\mathbf y^j = \mathcal N^e_{\theta^e} (\mathbf h^j) \approx \sum \limits_{i=1}^{k_{max}} \alpha_i^j \mathbf U_i, 
\end{equation}
where $(\alpha_1, ..., \alpha_{k_{max}})$ represents the $r$-latent roughness descriptors, enabling to represent any roughness profile, as well as generating many other roughness profiles by selecting values of $(\alpha_1, ..., \alpha_{k_{max}})$ different to the ones associated with the $P$ data profiles. The fact of operating into a linear latent vector space contribute to the procedure robustness with respect to the use of usual autoencoders.

However, until now, extracted modes $\mathbf U_i$, $i=1, ..., {k_{max}}$ and associated reduced descriptors $\alpha_i$, $i=1, ..., {k_{max}}$ were extracted with the only aim of representing the roughness. However, one could be interested in extracting the descriptors such that, other than representing the roughness, it or they should enable to classify the profiles for example into the different classes mentioned above. 

For that purpose a classifier operating on the reduced coordinates $\mathcal N^c_{\theta^c}$ is added to the RRAE architecture to associate to each latent representation $(\alpha_1, ..., \alpha_{k_{max}})$ its class $c \in \{ 1, ..., 12 \}$, that is 

\begin{equation}
c = \mathcal{N}^c_{\theta^c} \left( \sum_{i=1}^{k_{\max}} \alpha_i \mathbf{U}_i \right)  
\end{equation}

Thus, from the existing data ($\mathbf h^j, c^j$), $c^j$ being the class of profile $\mathbf h^j$, the training process operates to provide simultaneously the trainable parameters of the encoder, decoder and classifier neural network, while enforcing the linear vector latent space structure.

For this application, the encoder uses two one-dimensional convolutional blocks followed by a fully connected layer projecting into a latent space of dimension $d_\ell=200$. 
The first block applies 16 filters (kernel size 5, stride 1, padding 2) with ReLU activation and max-pooling, halving the spatial resolution. The second block increases channels to 32 (kernel size 3, stride 1, padding 1) with ReLU and max-pooling. Flattened feature maps are linearly mapped to the latent vector, preserving local correlations while reducing resolution by a factor of four.

The decoder reconstructs the signal via two linear layers expanding the latent vector, reshaped into 32 feature maps, followed by two transposed convolutions ($32 \rightarrow 6 \rightarrow 1$) with ReLU activations, restoring the original resolution.

A parallel feature predictor infers the classification from the latent space using a five-layer MLP (each of size $d_\ell$) with ReLU activations, ending with a linear layer to the target dimension. This branch enforces the latent space to encode information relevant for both reconstruction and classification.

\subsection{Results}

As previously indicated, the classifier was trained from both

\begin{enumerate}

\item The roughness profiles themselves  $\mathbf h^j$

\item The micro-roughness $\muv^j$ by removing the macro-roughness from the profiles

\end{enumerate}

When, using the profiles, first case above, a single descriptor was enough for performing an excellent, almost perfect, classification, in a very systematic way.

\begin{figure}
\centerline{\includegraphics[width=6 cm]{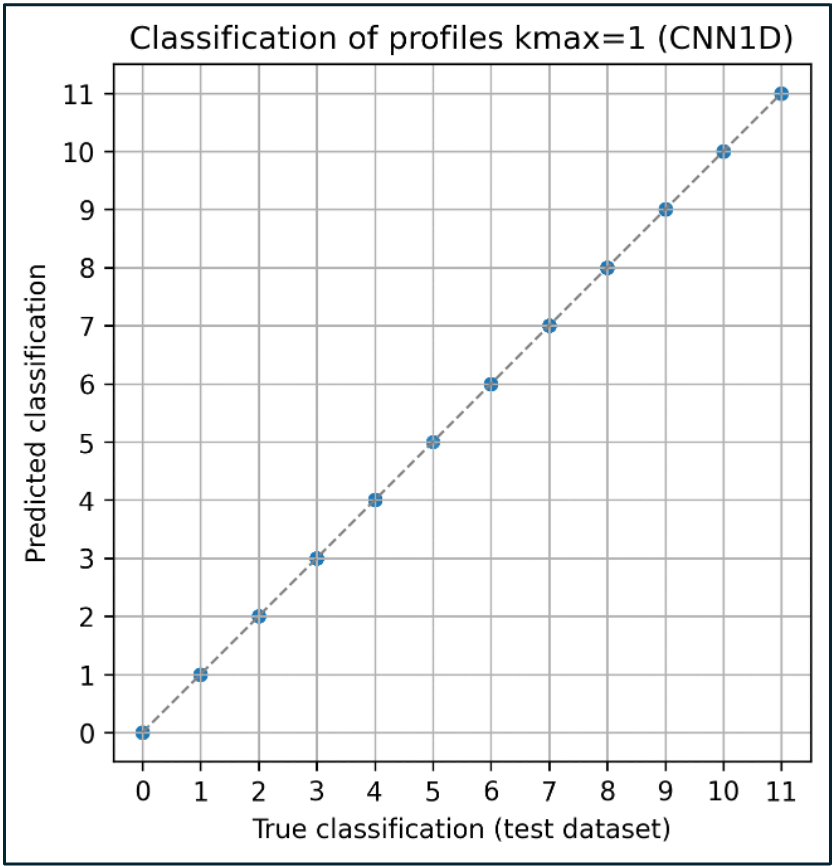}}
\caption{Classification accuracy when extracting a single descriptor for the measured profiles for classification purposes}
\label{ClasM}
\end{figure}

It is important to mention that $90 \%$ the available data was employed for training the RRAE architecture, and the remaining part for testing the accuracy of the proposed classification procedure. Figure~\ref{ClasM} represents the classification accuracy related to all the data, training and test. All data points lie on the $y=x$ axis indicating a perfect match between prediction and reference class value.

When considering only micro-roughness, we were interested in knowing if this micro-roughness contains descriptors or it is no more than a noise unable to inform on the tape origin (provider). For improving the neural capabilities, the encoder and decoder employed $1D$ convolution to better extract potential patterns involved within the micro-roughness profiles.

\begin{figure}
\centering
\includegraphics[width=5 cm]{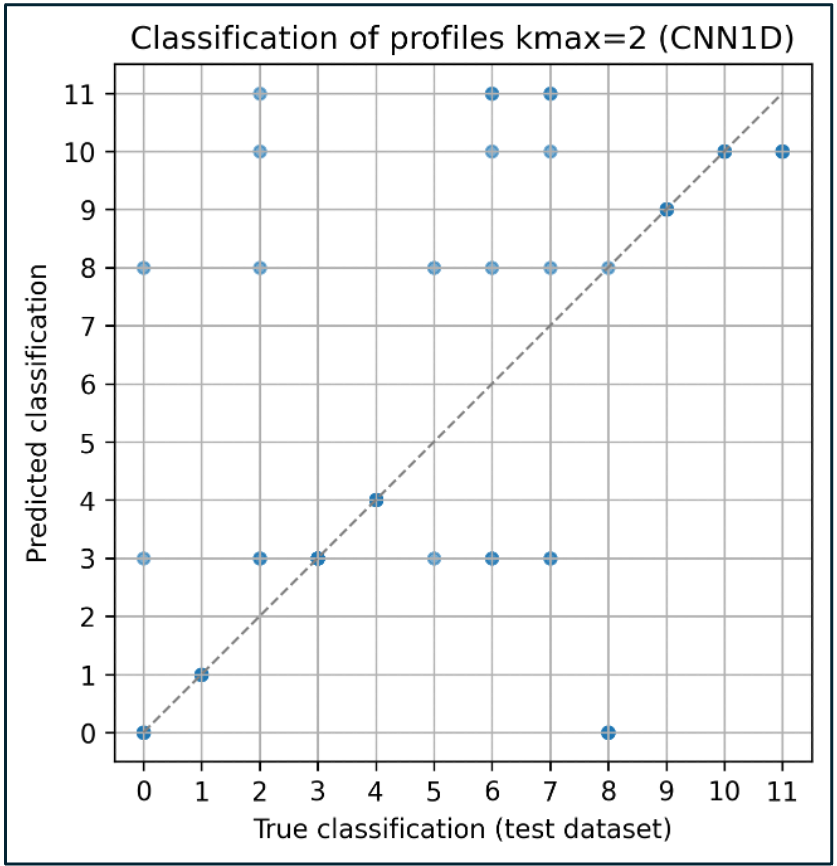}
\includegraphics[width=5 cm]{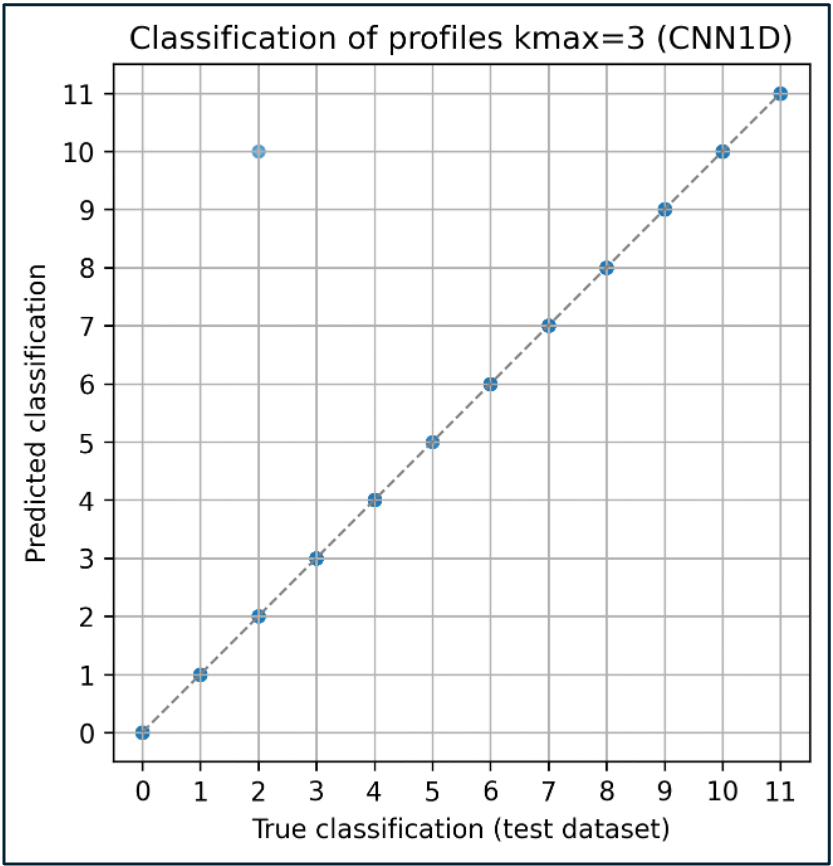}
\includegraphics[width=5 cm]{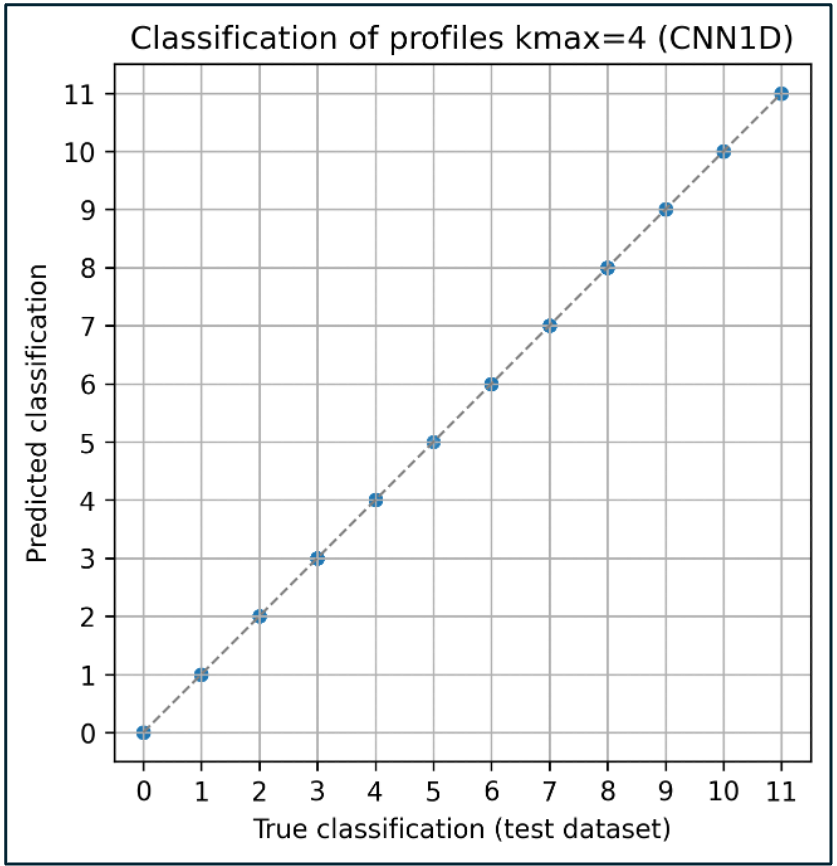}
\caption{Classification accuracy when classifying from the microscopic roughness for an increasing number of latent modes: $k_{max}=2$ (left), $k_{max}=3$ (center) and $k_{max}=4$ (right)}
\label{Clasm}
\end{figure}

Figure~\ref{Clasm} proves that four hidden descriptors enable classifying almost perfectly on the test data set as all data points sit on the $y=x$ axis. The same accuracy is also obtained on the training data set. The physical meaning of these four descriptors remains evasive, but the proposed rationale proves that the micro-roughness contains information beyond a simple noise, and that increasing the number of descriptors from one (when using the real profiles) to four (when dealing with micro-roughness only) enables performing an almost perfect classification.

\section{Rough interfaces consolidation}\label{sec:consolidation}

Section~\ref{sec:class} demonstrated how the Rank Reduction AutoEncoder (RRAE) can extract a small set of relevant latent descriptors that are effective for classifying roughness profiles. The present section extends this methodology to identify descriptors that allow the prediction of the evolution of the degree of intimate contact for a given micro-roughness profile $\muv(x)$.

Because the classification task and the intimate contact are not fully independent, a unified RRAE architecture may also be considered, in which both objectives are learned simultaneously.

\subsection{Data generation}

\begin{figure}[!h]
\centering
\includegraphics[width=\textwidth]{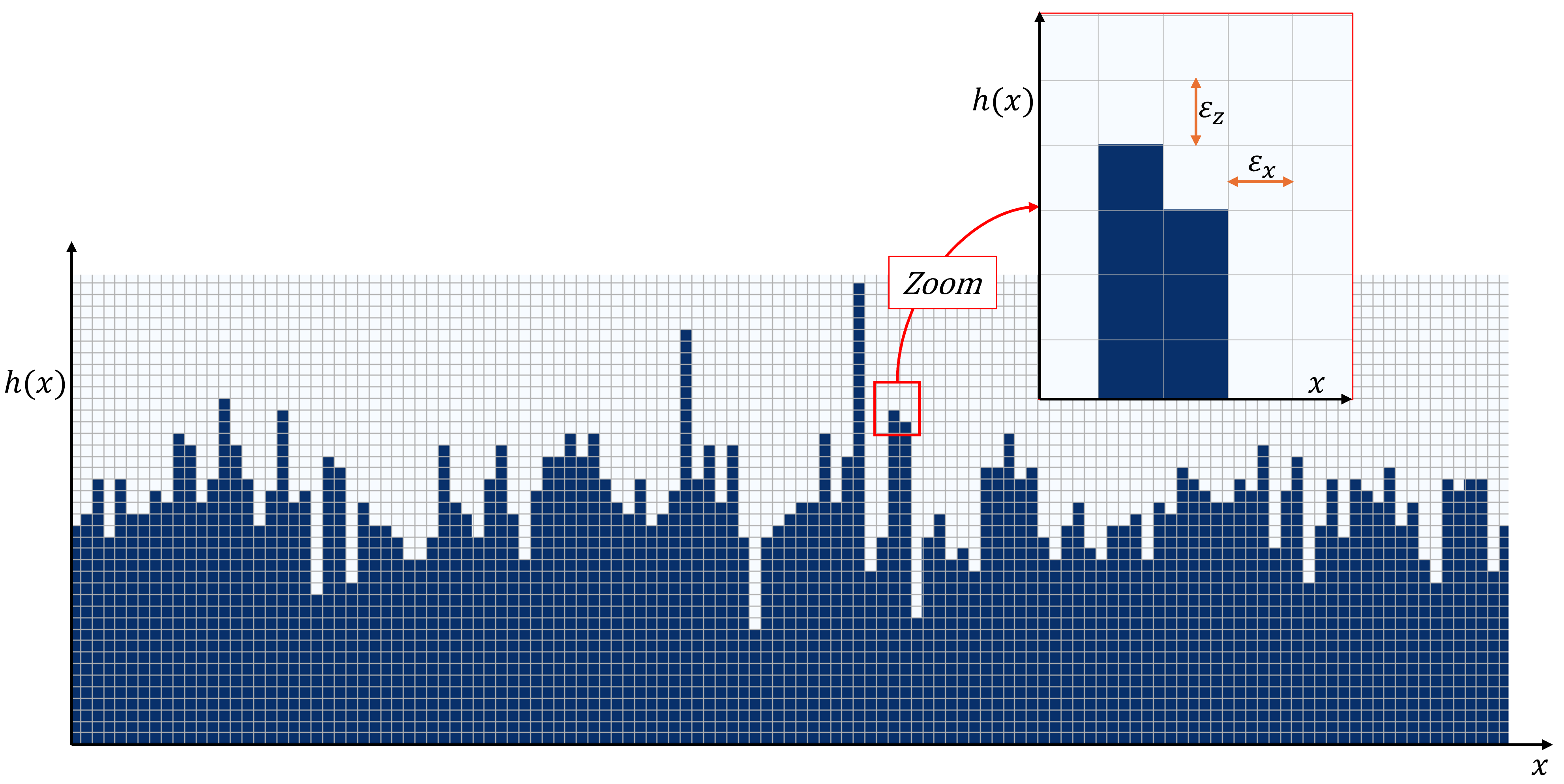}
\caption{Illustrating a portion of a discretized micro-roughness profile: blue elements represent polymer material while clear element represent air.}
\label{fig:discretization}
\end{figure}

The intimate contact is computed using the cellular automaton introduced in Section~\ref{sec:sim}.  The transverse roughness profile is discretized as illustrated in Fig.~\ref{fig:discretization}.  Assuming a constant compressing plate advancing velocity, material cells (blue) progressively fill neighboring air cells (grey).

At each time step, the degree of intimate contact (DIC) is defined as the ratio between the contact length and the total tape width (see Section~\ref{sec:sim}).  This quantity increases with time and approaches perfect contact ($\mathrm{DIC}=1$) if sufficient time is allowed.

The accuracy of the DIC evolution depends on the spatial discretization. Two resolutions are distinguished in Fig.~\ref{fig:discretization}: The horizontal discretization along the $x$-direction, $\epsilon_x$, and the vertical discretization, $\epsilon_z$.

The value of $\epsilon_x$ is imposed by the measurement system. It corresponds to the sampling distance and can be enlarged by averaging neighboring points or reduced by interpolation.  For the experimental profiles considered here, $\epsilon_x  \approx 3~\mu m$ for a tape width of about 3~mm. This value is dictated by the 3D non-contact profilometer and is therefore kept unchanged.

On the other hand, $\epsilon_z$ is a numerical parameter that strongly influences the simulation accuracy. As in the Finite Element Method, smaller values improve the resolution but increase the computational cost. A convergence analysis indicates that the DIC evolution becomes insensitive to the discretization for $\epsilon_z \leq 0.1~\mu m$; further refinement leads to negligible accuracy improvement while significantly increasing runtime.

\begin{figure}[!h]
\centering
\includegraphics[width=0.8\textwidth]{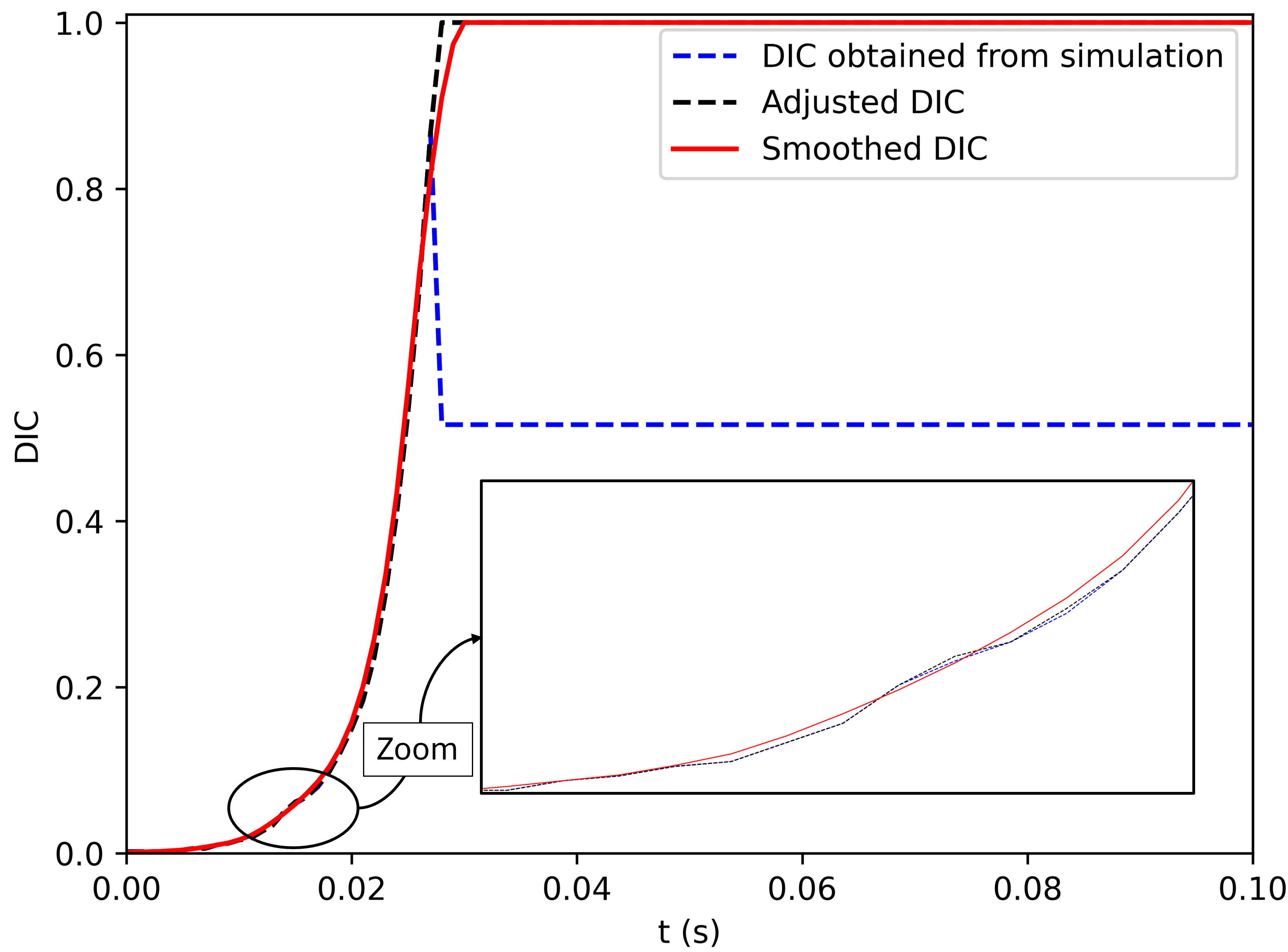}
\caption{DIC evolution curve for a given micro-roughness profile.}
\label{fig:ProcessedDic}
\end{figure}

A numerical artifact induced by the spatial discretization, and in particular by the choice of $\epsilon_z$, concerns the converged value of the DIC. 
As illustrated in Fig.~\ref{fig:ProcessedDic} for a representative micro-roughness profile over 100 time steps, 
the DIC increases with time until it may abruptly drop to a final plateau.  This last value is non-physical: It corresponds to the number of material cells that remain in contact with the compressing plate, $N_c$, once no air cells are available anymore. Because the cellular automaton does not allow further compaction of the tape cells, it can be mathematically demonstrated that the asymptotic value becomes
\[
\mbox{DIC}(t_{\infty}) = \frac{N_c}{N_w} - \left\lfloor \frac{N_c}{N_w} \right\rfloor,
\]
where $N_w$ denotes the number of cells along the width and $\lfloor \cdot \rfloor$ is the floor operator.

This numerical artifact disappears when $\epsilon_z \rightarrow 0$, which is practically impossible to achieve. Therefore, to remain consistent with the underlying physics, this artificial value is replaced by $1$, leading to the corrected DIC curve shown in Fig.~\ref{fig:ProcessedDic}.  

Furthermore, to reduce the sensitivity of the response to the choice of $\epsilon_z$, the DIC history is smoothed using a moving average with a window of five time steps. This operation removes small oscillations caused by discrete contact jumps and softens the sharp transition observed when the DIC approaches unity.

These post-processing steps improve the data quality by limiting numerical artifacts, allowing the machine learning models to capture the physics rather than fit discretization noise.

Finally, to reduce the computational burden, each simulation is stopped as soon as the converged DIC value is reached. For the 1011 investigated cases, this strategy resulted in a total runtime of approximately 24~hours on a system equipped with a 13th~Gen Intel(R) Core(TM) i9-13950HX processor (2.20~GHz) and 32~GB of RAM. A maximum of 352 time steps was sufficient to guarantee convergence for all considered micro-roughness profiles.

\subsection{Modeling approach}

\begin{figure}[!h]
\centering
\includegraphics[width=0.8\textwidth]{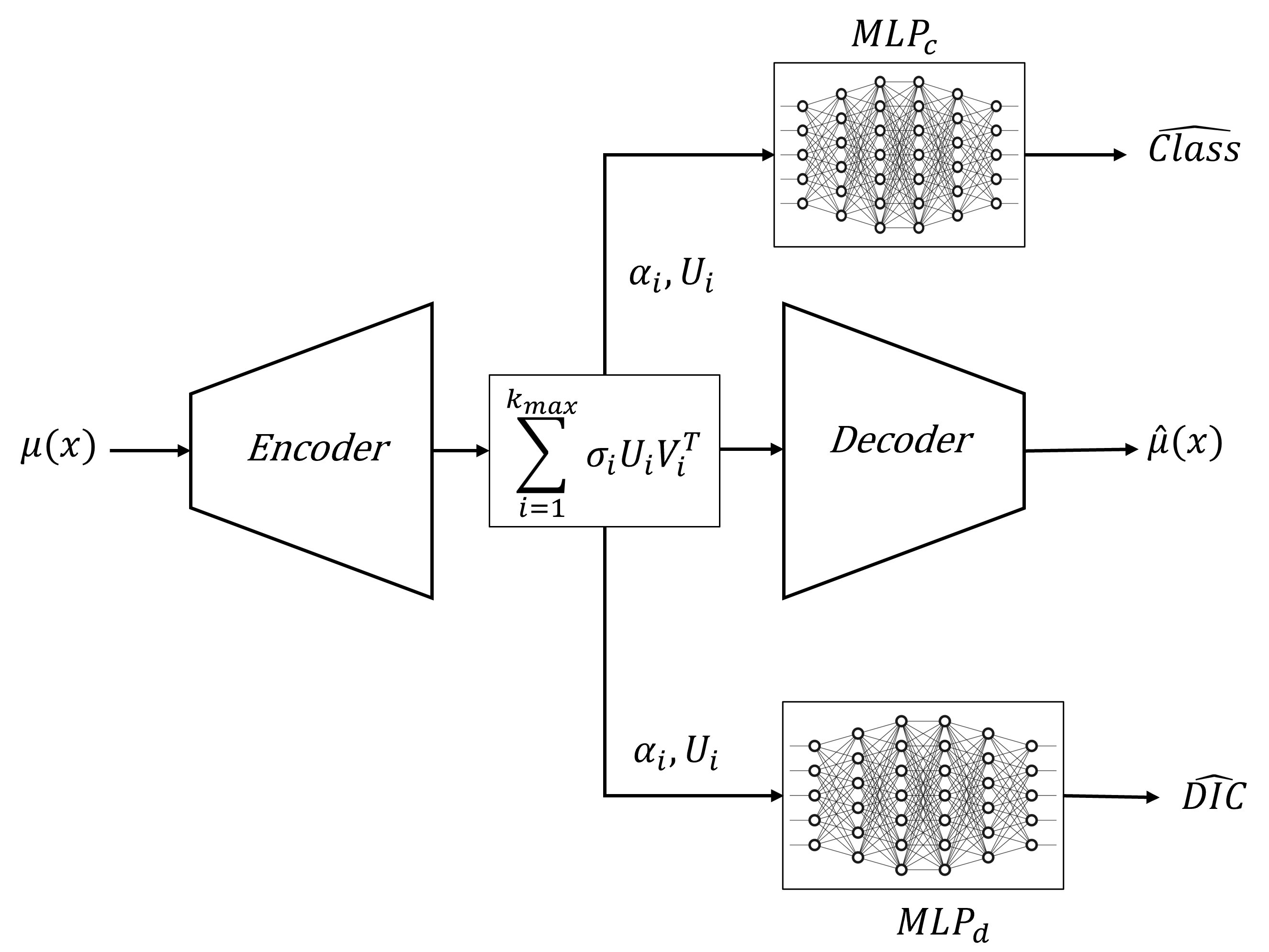}
\caption{Rank Reduction AutoEncoder architecture for simultaneous inference of DIC and classification.}
\label{fig:RRAE2MLP}
\end{figure}

In this section, the RRAE introduced in Section~\ref{sec:class} is extended to predict the DIC evolution curve simultaneously with the classification of micro-roughness profiles.  The encoder and decoder architectures remain unchanged, while two Multi-Layer Perceptrons (MLPs) are connected to the latent space of the autoencoder, as illustrated in Fig.~\ref{fig:RRAE2MLP}. The first MLP, $MLP_c$, predicts the classification, whereas the second, $MLP_d$, predicts the DIC curve. While achieving both predictions within a single architecture can be challenging for standard machine learning methods, the RRAE handles it naturally.

Both MLPs consist of six layers with a number of neurons equal to the latent space dimension (here, 200), except for the final layer, which matches the output size (12 for classification, 352 for DIC). The RRAE is relatively insensitive to the exact MLP design, allowing for larger networks without overfitting. However, the choice of $k_{\max}$ remains critical and should be explored.

The overall loss function is defined as:
\begin{equation}
L = \omega_r L_r + \omega_c L_c + \omega_{d} L_{DIC},
\end{equation}
where $\omega_r$, $\omega_c$, and $\omega_{d}$ are weighting factors. 
$L_r$ corresponds to the reconstruction loss of the micro-roughness profile, $L_c$ to the classification loss, and $L_{DIC}$ to the DIC prediction loss.  

Each loss is computed as a relative L2-norm:
\begin{equation}
\mathrm{loss} = \frac{\|\hat{x} - x\|_2}{\|x\|_2},
\end{equation}
where $\hat{x}$ is the predicted quantity and $x$ the reference. This formulation measures the inference error as a percentage relative to the L2-norm of the target.

In order to give all three outputs the same importance, equal weight are used thereafter ($\omega_r=\omega_c=\omega_{d}=1$).

To ensure that the reconstruction loss $L_r$ properly reflects discrepancies between predicted and reference roughness profiles, an additional normalization step was introduced. Each micro-roughness signal $\muv$ was first normalized $\muv_{norm} \in[0,1]$ using:
\begin{equation}
\muv_{norm}^i = \frac{\muv_i - \min\limits_{j}(\muv_j)}{\max\limits_{j}(\muv_j) - \min\limits_{j}(\muv_j)},
\end{equation}
for $i \in [1, P]$, where $P$ is the number of micro-roughness profiles.
However, despite this normalization, the signals remain highly concentrated around their mean value $\tilde{\muv}_{norm}$. In the original data, roughness amplitudes range approximately from -32 to 35~$\mu$m, but with a small standard deviation $\sigma(\muv)=1.3~\mu$m. After normalization, this reduces to $\sigma(\muv_{norm})=0.018$, meaning that most values lie within $\tilde{\muv}_{norm}\pm0.056$. Consequently, the relative $L_2$-error corresponds to variations below $5\%$, causing rapid loss saturation and limited gradient sensitivity during training.

To enhance contrast between profiles, the normalized signal was further standardized as:
\begin{equation}
\muv_{\mathrm{mod}}^i = \frac{\muv_{norm}^i - \tilde{\muv}_{norm}}{6\,\sigma(\muv_{norm}))}, 
\end{equation}
for $i \in [1, n]$. This transformation amplifies relative differences between profiles, increasing the effective $L_2$-variation to above $25\%$ and improving gradient-based optimization. Without this rescaling, the reconstruction accuracy reported in the next section could not be achieved.

The dataset was divided into 911 samples for training and 100 for testing, the latter being randomly selected across all classes. Training was performed for $30\,000$ epochs with an initial learning rate of $10^{-3}$, reduced by a factor of ten every $10\,000$ epochs. 
On the hardware described previously, using an RTX4080 GPU with 12~GB of memory, the complete training required less than ten minutes for which the loss function reached a plateau of around $2\%$ error value.

\subsection{Results}

Different values of $k_{max}$ were explored, as this parameter is central to the RRAE architecture. The results indicate that $k_{max}=5$ is sufficient to accurately reproduce surface roughness, perform classification, and predict the DIC evolution. Increasing $k_{max}$ to 6 or higher ($k_{max}\geq6$) provides only marginal improvements. This suggests that five latent features are sufficient to capture the essential complexity of the interface roughness for both classification and DIC estimation tasks.

\begin{figure}[!h]
\centering
\includegraphics[width=0.8\textwidth]{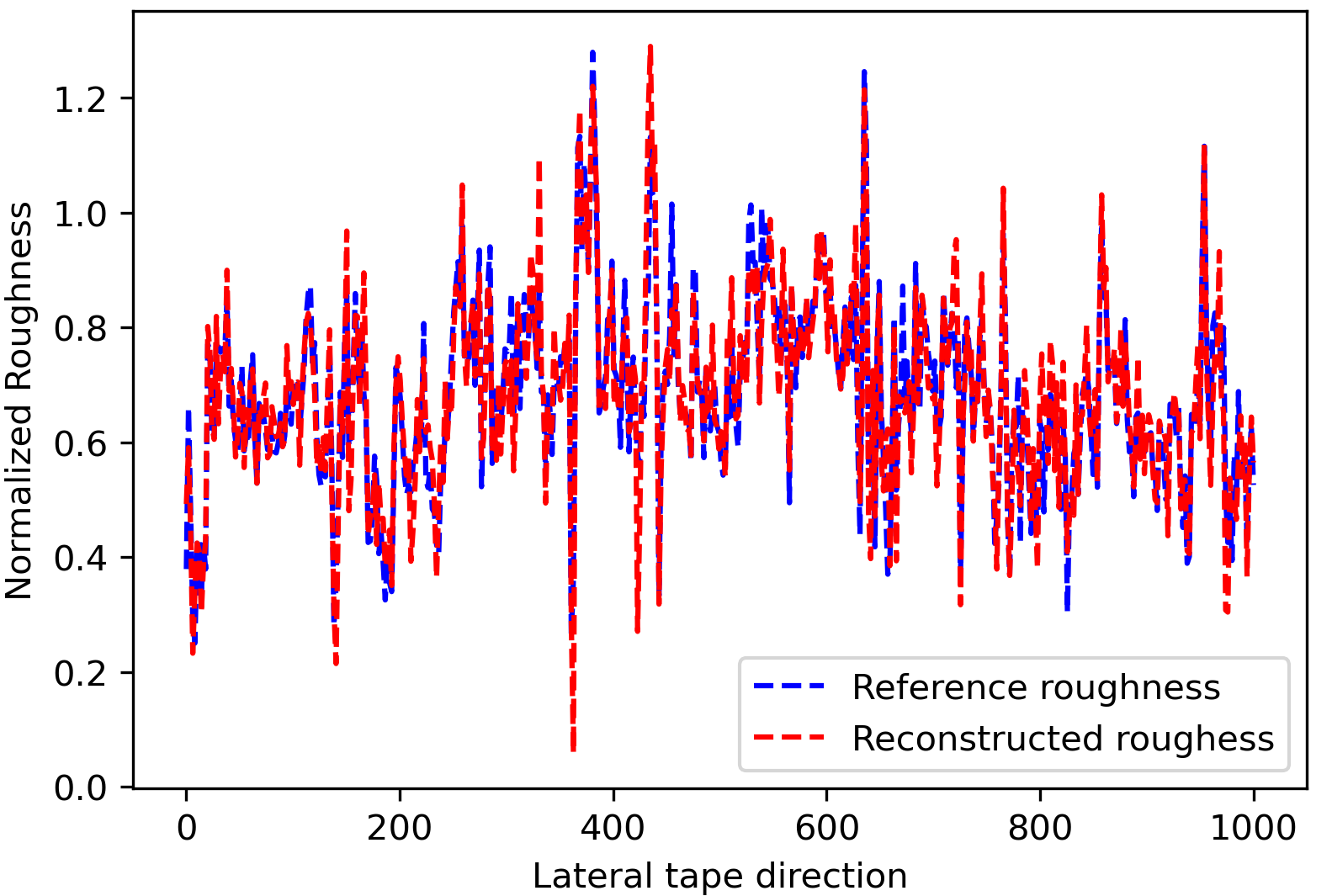}
\caption{Roughness reconstruction for a representative test case with $k_{max}=5$.}
\label{fig:RRAERug}
\end{figure}

This choice also ensures accurate roughness reconstruction. For $k_{max}=5$, the average reconstruction error (L2-norm) across test profiles is approximately $5\%$, as illustrated in Fig.~\ref{fig:RRAERug}. Errors increase significantly for $k_{max}<4$, while $k_{max}=4$ already limits the reconstruction error below $10\%$. Further increasing $k_{max}$ beyond 6 does not yield meaningful improvements.

\begin{figure}[!h]
\centering
        \begin{subfigure}{0.48\textwidth}
            \includegraphics[width=\textwidth]{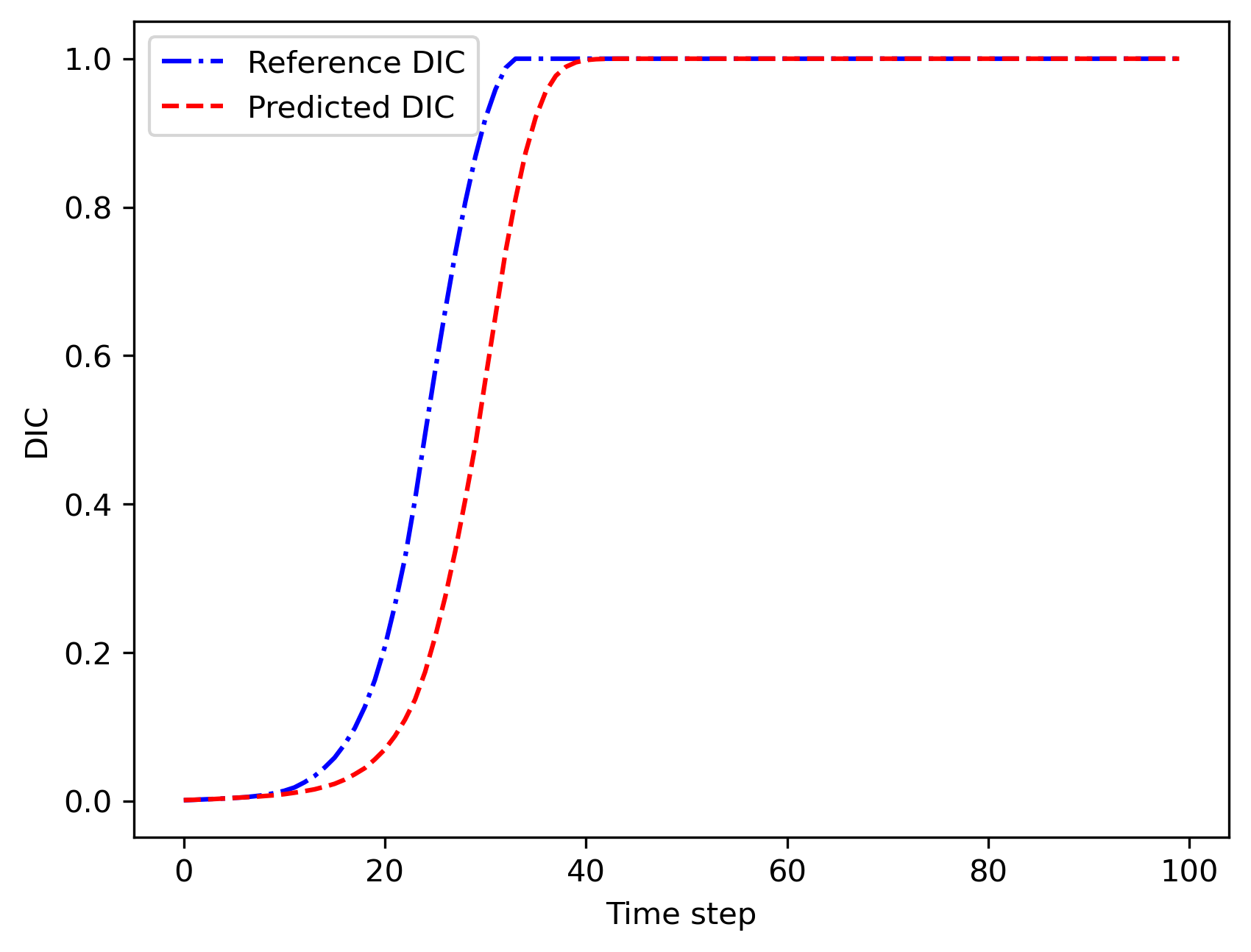}
            \caption{Average prediction accuracy}
            \label{fig:RRAEDicOut_3}
        \end{subfigure}  
    \hfill
        \begin{subfigure}{0.48\textwidth}
            \includegraphics[width=\textwidth]{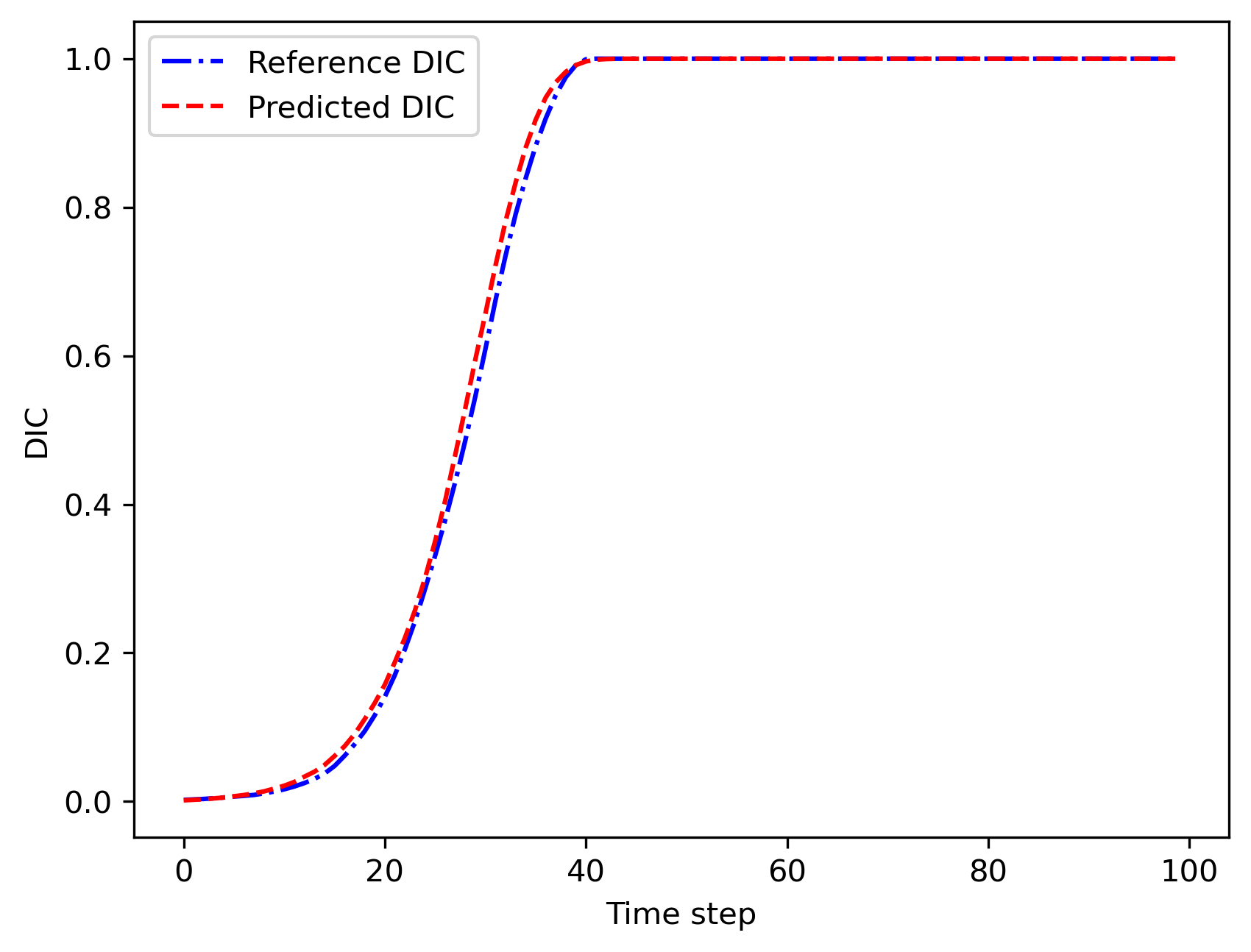}
            \caption{High prediction accuracy}
            \label{fig:DIC_Test_Kmax5case3}
        \end{subfigure}  
        \vfill
        \begin{subfigure}{0.48\textwidth}
            \includegraphics[width=\textwidth]{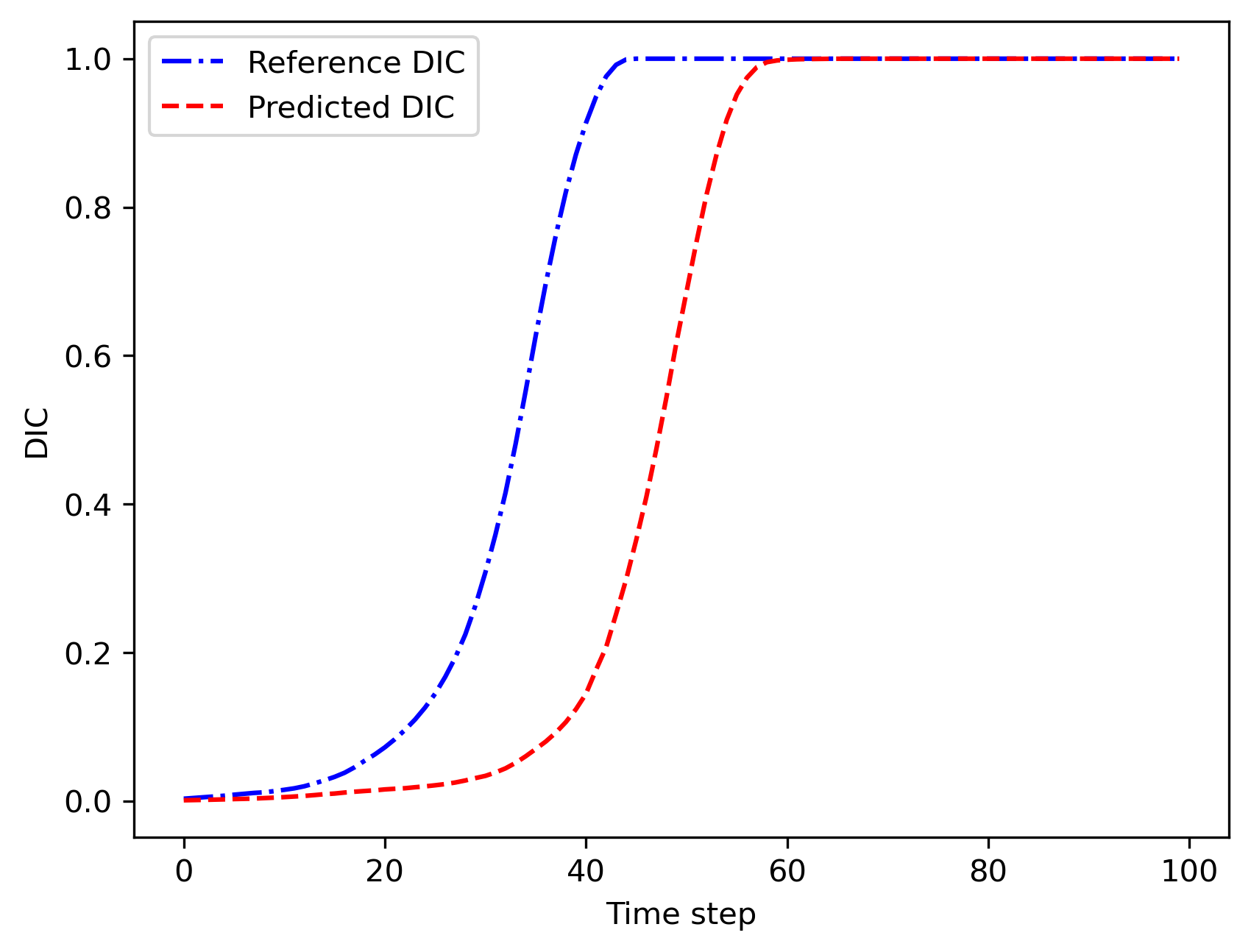}
            \caption{Lower prediction accuracy}
            \label{fig:DIC_Test_Kmax5case19}
        \end{subfigure} 
        \hfill
        \begin{subfigure}{0.48\textwidth}
            \includegraphics[width=\textwidth]{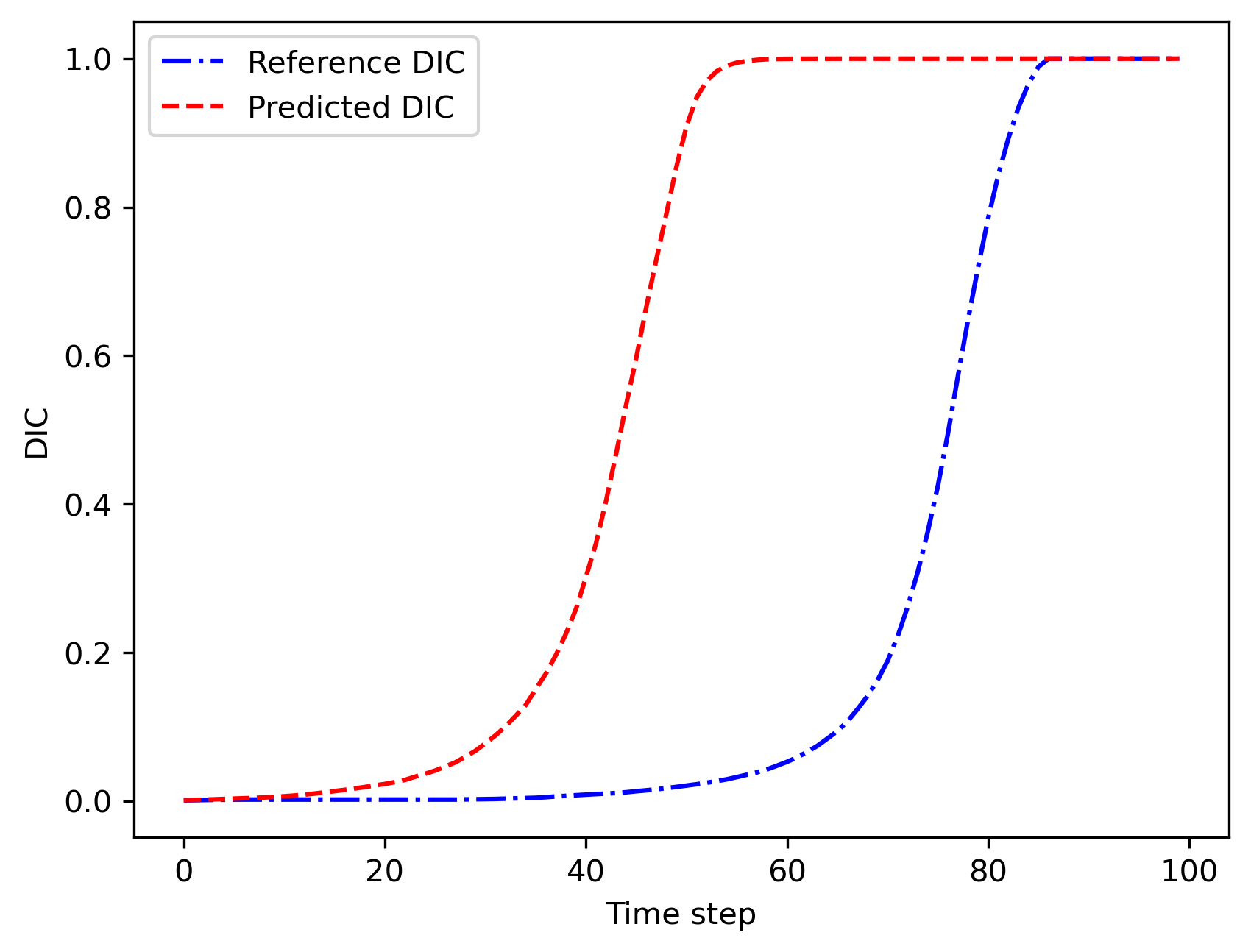}
            \caption{Highest error case}
            \label{fig:DIC_Test_Kmax5case48}
        \end{subfigure}   
\caption{Prediction of the DIC evolution curve using the RRAE for four representative micro-roughness profiles. Figures zoomed on the first 100 time steps, above which DIC is constant and equal to 1.}
\label{fig:RRAEDicOut}
\end{figure}

Figure~\ref{fig:RRAEDicOut} presents representative DIC predictions obtained with the RRAE ($k_{max}=5$) for four micro-roughness profiles from the test set. The model captures the overall temporal evolution of the DIC response with good agreement. In the best cases (Fig.~\ref{fig:DIC_Test_Kmax5case3}), the prediction nearly overlaps the reference curve. Figure~\ref{fig:RRAEDicOut_3} illustrates an average case, while Fig.~\ref{fig:DIC_Test_Kmax5case19} shows a lower-accuracy example, where a small horizontal shift appears along the evolution. Figure~\ref{fig:DIC_Test_Kmax5case48} corresponds to one of the few outliers exhibiting the largest prediction error. Overall, discrepancies remain limited and primarily affect a small subset of challenging profiles.

For all 100 test cases, the prediction error manifests predominantly as a horizontal shift between the reference and predicted DIC curves, without significant distortion of their shape. The error is therefore quantified by computing the area corresponding to this shift and normalizing it by the area under the reference curve. This normalized metric is denoted $\delta_{DIC}$ and expressed as a percentage.

\begin{figure}[!h]
\centering
        \begin{subfigure}{0.48\textwidth}
            \includegraphics[width=\textwidth]{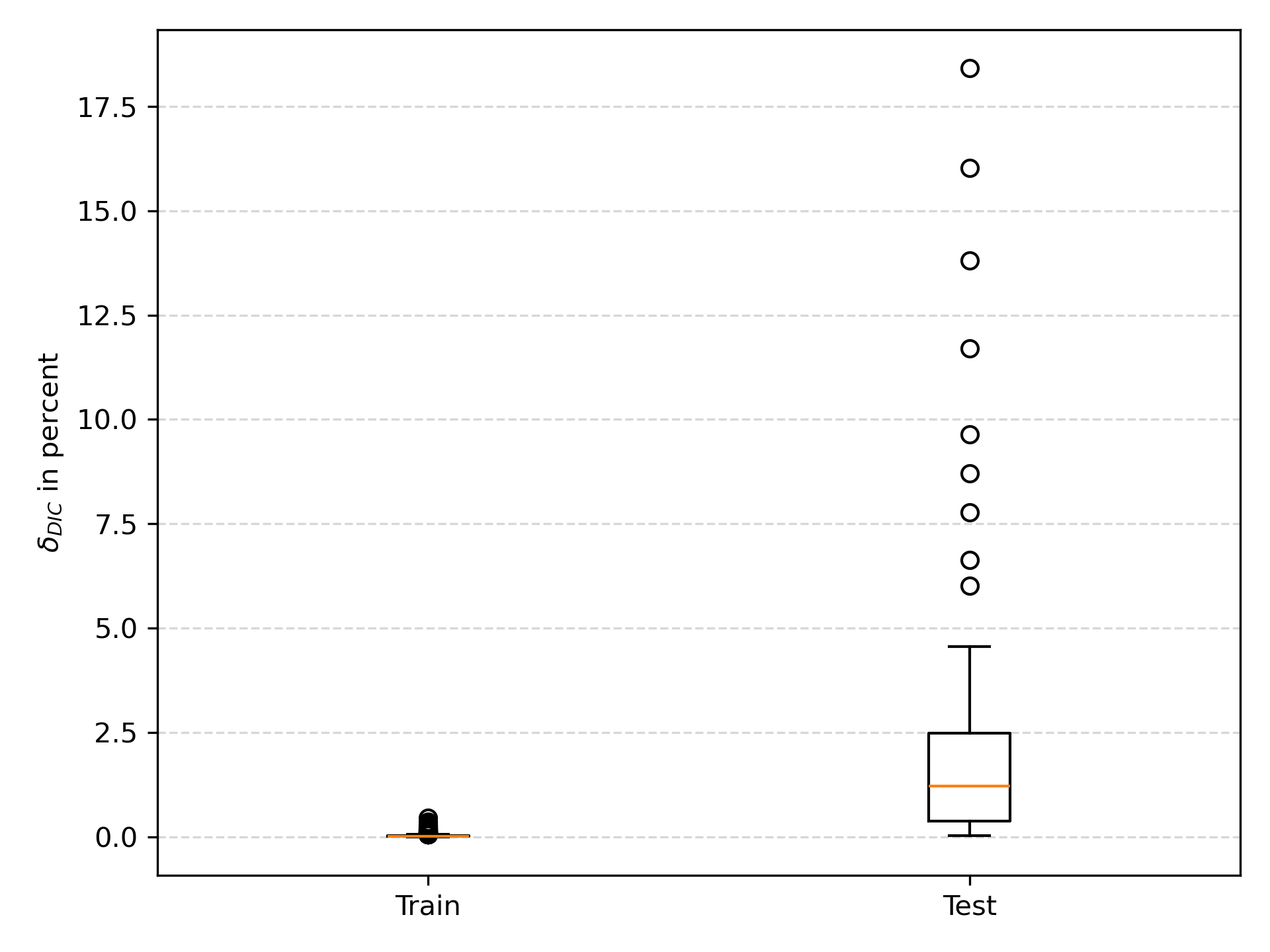}
            \caption{Boxplot distribution of $\delta_{DIC}$}
            \label{fig:RRAEDicOut_Box5}
        \end{subfigure}  
    \hfill
        \begin{subfigure}{0.48\textwidth}
            \includegraphics[width=\textwidth]{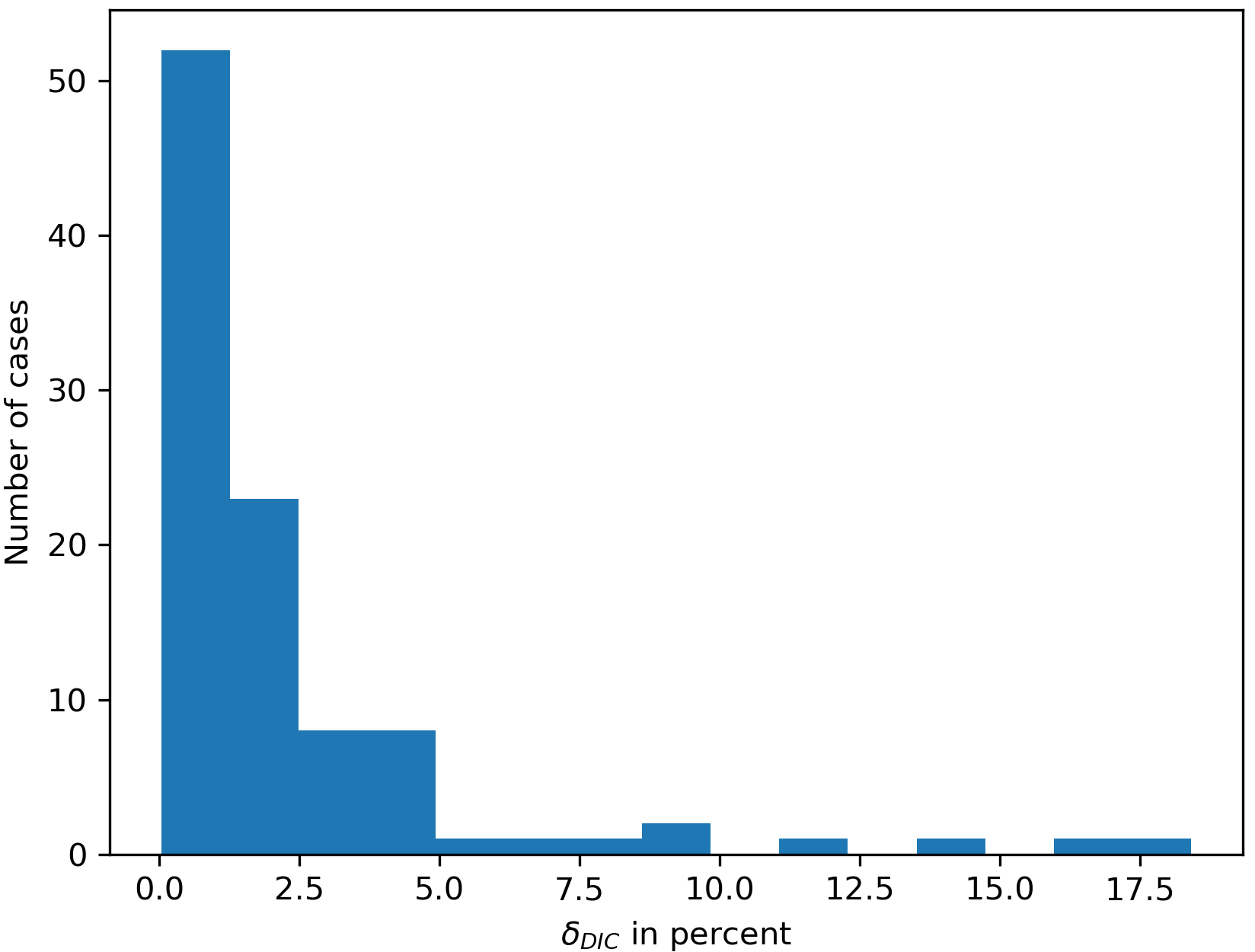}
            \caption{Histogram of test errors}
            \label{fig:RRAEDicOut_Histo5}
        \end{subfigure}  
\caption{Accuracy statistics for the DIC prediction with $k_{max}=5$.}
\label{fig:RRAEDicOut_Stat}
\end{figure}

Figure~\ref{fig:RRAEDicOut_Box5} shows the boxplot distribution of $\delta_{DIC}$ for both training and test datasets, while Fig.~\ref{fig:RRAEDicOut_Histo5} presents the histogram of test errors. The average test error is approximately $2\%$, with only three outliers exceeding $10\%$. The largest outlier corresponds to Fig.~\ref{fig:DIC_Test_Kmax5case48}. These cases are likely due to the limited size of the training dataset, meaning that certain roughness patterns in the test set are insufficiently represented during training.

For the training dataset, the average error remains below $1\%$, and all 911 samples exhibit errors lower than $5\%$. Although this gap between training and test errors could suggest mild overfitting, applying early stopping reduced test accuracy. Similarly, introducing dropout or modifying the encoder, decoder, or MLP components did not significantly improve generalization performance.

\begin{figure}[!h]
\centering
        \begin{subfigure}{0.48\textwidth}
            \includegraphics[width=\textwidth]{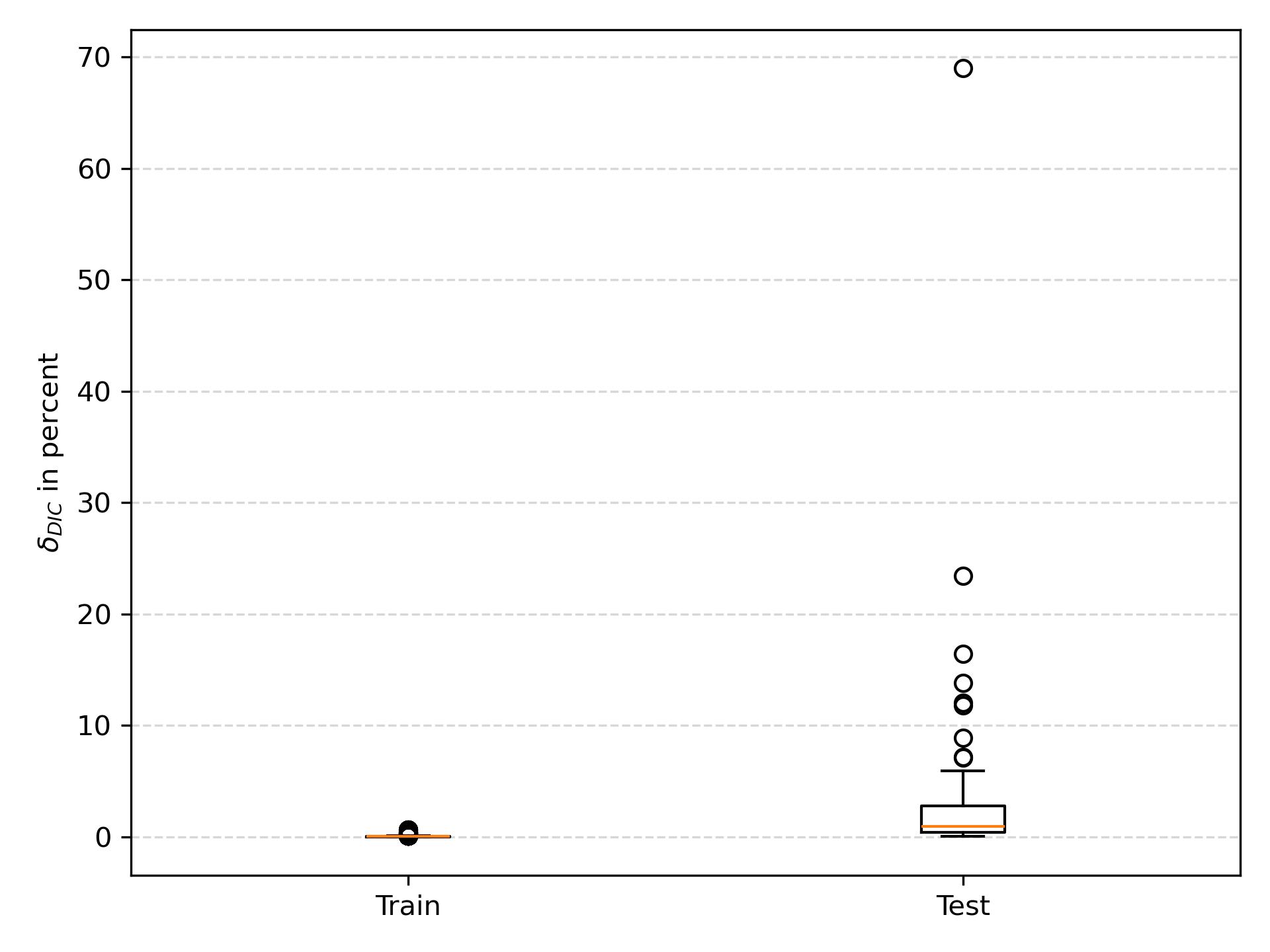}
            \caption{Boxplot distribution of $\delta_{DIC}$}
            \label{fig:RRAEDicOut_Box4}
        \end{subfigure}  
    \hfill
        \begin{subfigure}{0.48\textwidth}
            \includegraphics[width=\textwidth]{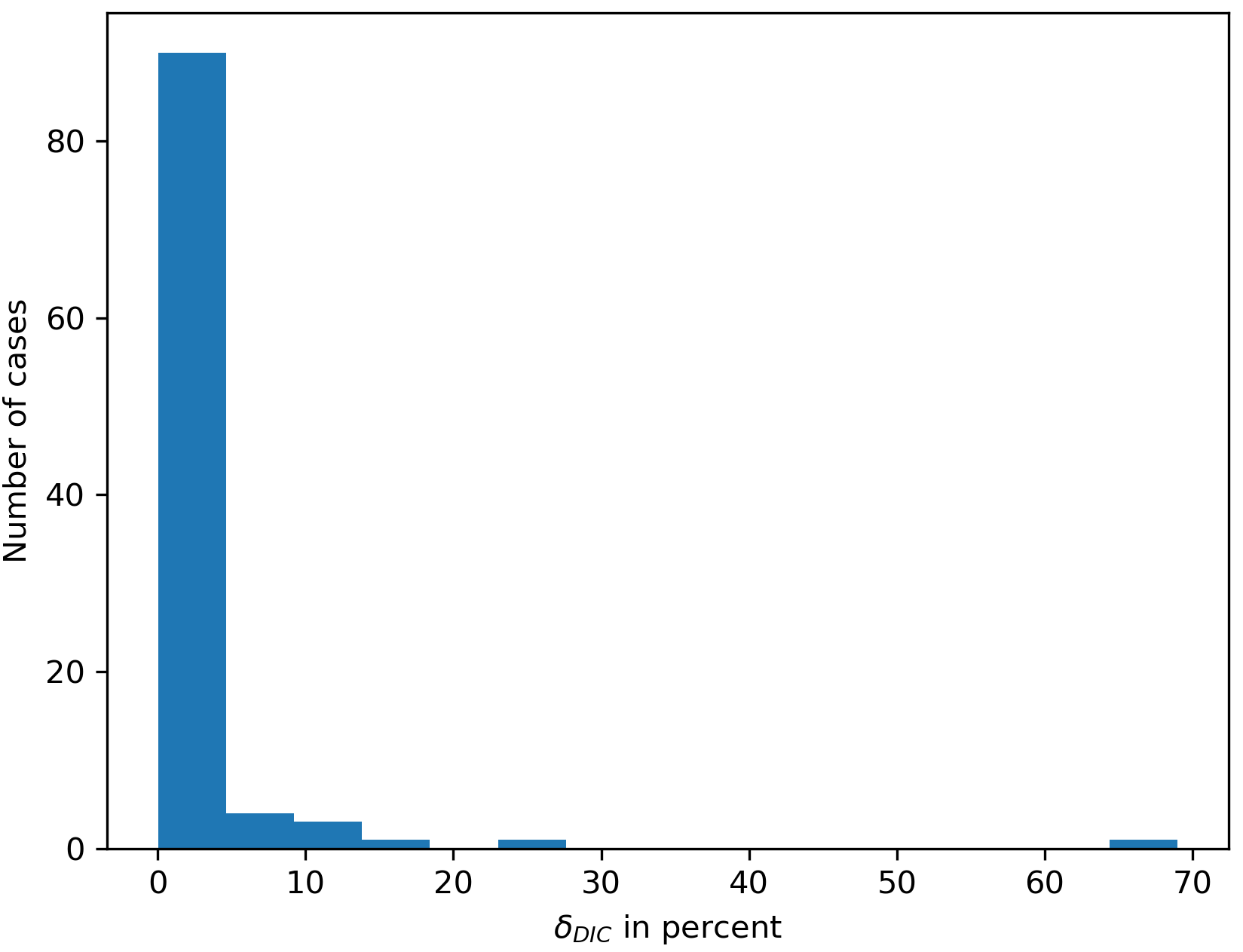}
            \caption{Histogram of test errors}
            \label{fig:RRAEDicOut_Histo4}
        \end{subfigure}  
\caption{Accuracy statistics for the DIC prediction with $k_{max}=4$.}
\label{fig:RRAEDicOut_Stat4}
\end{figure}

\begin{figure}[!h]
\centering
\includegraphics[width=0.4\textwidth]{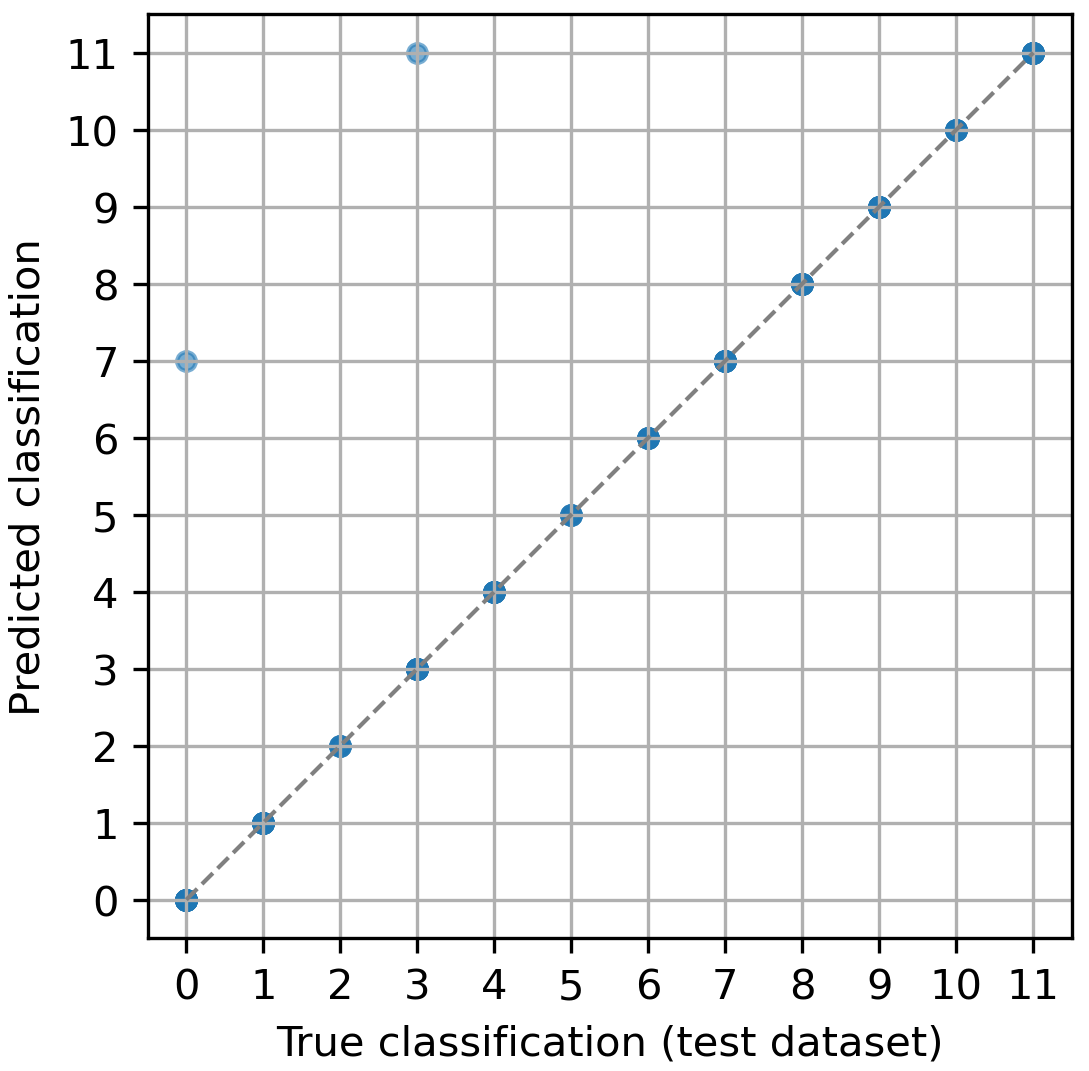}
\caption{Classification results for $k_{max}=4$ using the RRAE architecture of Fig.~\ref{fig:RRAE2MLP}.}
\label{fig:ClassK4}
\end{figure}

The cumulative $\delta_{DIC}$ over all test cases is approximately 221 for $k_{max}=5$, compared to about 287 for $k_{max}=4$, corresponding to an error reduction of roughly $30\%$. The associated statistics for $k_{max}=4$ are shown in Fig.~\ref{fig:RRAEDicOut_Stat4}. Although results with $k_{max}=4$ remain acceptable (except for a few outliers exceeding $20\%$), classification performance deteriorates, with two incorrect predictions (Fig.~\ref{fig:ClassK4}). 

This observation indicates that the features governing DIC prediction and those governing classification are not strictly identical. While some latent features are shared between tasks, others are task-specific. Consequently, five latent features provide a suitable compromise to ensure accurate performance for both classification and DIC prediction.

In summary, five latent features are sufficient to describe the roughness profile with adequate fidelity to enable both tape classification and compaction quality prediction. 

\subsection{Decoupled Feature Learning: An Extended RRAE Architecture}

The previous analysis suggested that four latent features were insufficient to simultaneously achieve high classification accuracy and precise DIC prediction. This observation could indicate that classification and compaction rely on partially distinct hidden features. However, an alternative explanation is that the original architecture may not optimally disentangle the latent representations required for both tasks.
\begin{figure}[!h]
\centering
\includegraphics[width=\textwidth]{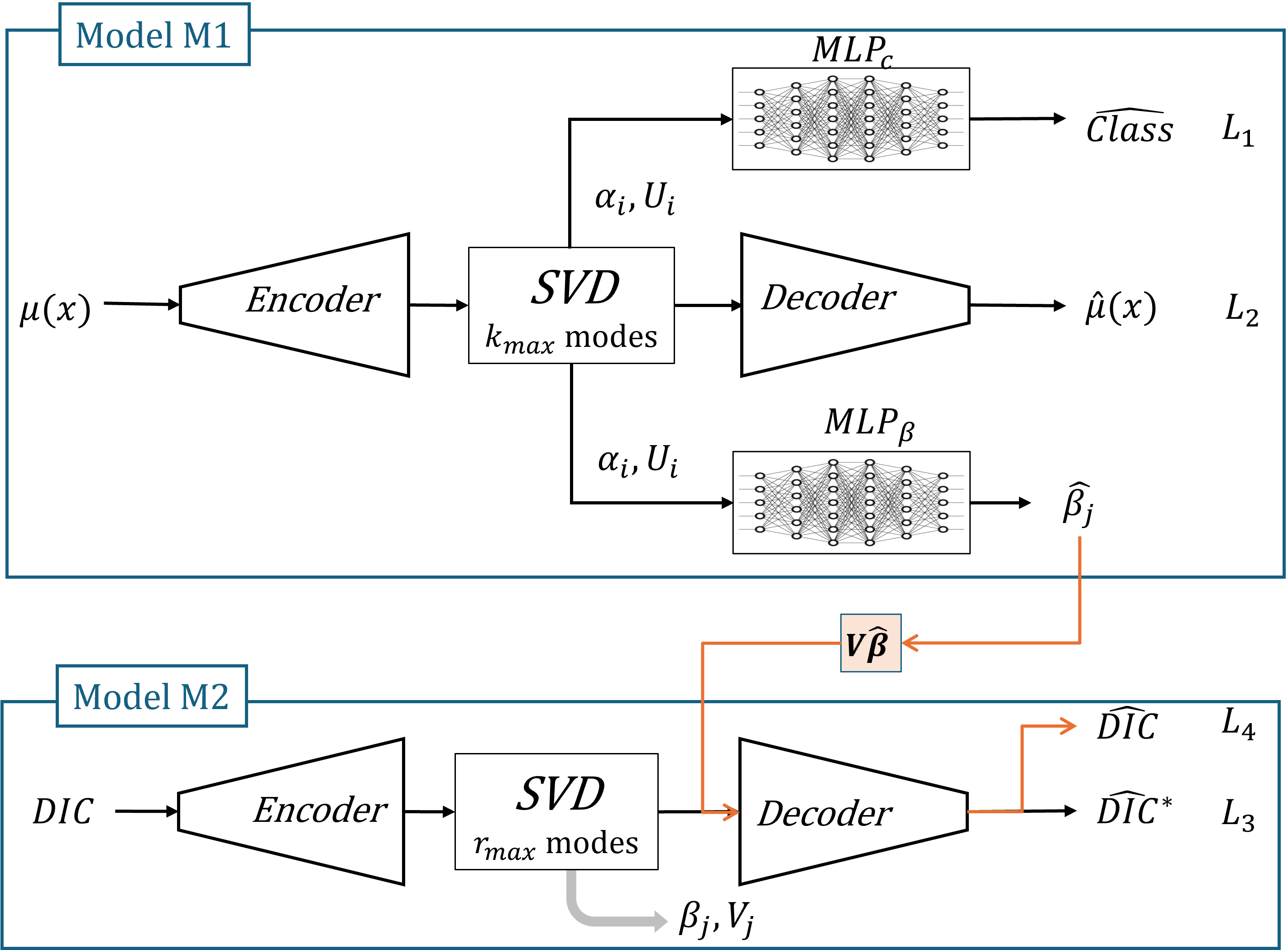}
\caption{Extended architecture with decoupled latent representations: model $M_1$ extracts roughness features and predicts modal coefficients $\hat{\mathbf{\beta}}$, while model $M_2$ reconstructs the DIC using a fixed SVD basis $\mathbf{V}$.}
\label{fig:Archi_AlphaBeta}
\end{figure}

To investigate this hypothesis, a more advanced architecture is proposed (Fig.~\ref{fig:Archi_AlphaBeta}). The idea is to decouple the discovery of roughness-related features from the intrinsic modal representation of the DIC response by introducing two jointly trained RRAE models.

Model $M_1$ corresponds to the original RRAE architecture (Fig.~\ref{fig:RRAE2MLP}), with one modification: the MLP branch previously used to predict the DIC ($MLP_d$) is replaced by a network denoted $MLP_\beta$. Instead of directly predicting the DIC curve, $MLP_\beta$ outputs a vector of reduced coefficients $\hat{\mathbf{\beta}} = \{\hat{\beta}_j\}$. Importantly, no activation function is applied after the final linear layer, since the coefficients $\beta_j$ are not constrained to a bounded interval (unlike the DIC, which lies in $[0,1]$).

Model $M_2$ is designed to independently learn the intrinsic reduced structure of the DIC response. At the SVD level, it outputs both the reduced coefficients $\mathbf{\beta}$ and a fixed basis $\mathbf{V}$, such that $\mathbf{V}\mathbf{\beta}$ reconstructs the latent representation feeding the decoder. The reconstructed DIC is denoted $\hat{DIC}^\star$. When trained separately, $M_2$ isolates the intrinsic dimensionality of the DIC evolution. The results confirm that only $r_{max}=3$ modes are sufficient to reconstruct the DIC with less than $5\%$ error for all training and testing profiles.

The two models are then trained simultaneously. During training, the predicted coefficients $\hat{\mathbf{\beta}}$ from $M_1$ are injected into the SVD layer of $M_2$ through the operation $\mathbf{V}\hat{\mathbf{\beta}}$, which feeds the decoder to produce $\hat{DIC}$. The modal basis $\mathbf{V}$ is obtained from SVD with $r_{max}=3$ and fixed after training. The joint training of both models requires a common loss function that can be written:
\begin{equation}
Loss = \Sigma_{i=1}^4 \omega_i L_i,
\end{equation}
where $L_i$ is the loss function related to the: 
\begin{itemize}
    \item tape classification: $L_1$ 
    \item reconstruction of the roughness profile: $L_2$
    \item reconstruction of the DIC ($\hat{DIC}^\star$): $L_3$ 
    \item prediction of the DIC ($\hat{DIC}$): $L_4$.
\end{itemize}
All loss functions are calculated according to the $L2$-norm, with $\omega_i$ for $i=1, \dots, 4$ is an associated weight. To increase the importance of the predicted features, the weight coefficients are chosen as follows: $\omega_1=\omega_4=2$ while $\omega_2=\omega_3=1$.

At inference, no prior DIC information is required. The micro-roughness profile $\mu(x)$ is input to $M_1$, yielding both the classification $\hat{Class}$ and the coefficients $\hat{\mathbf{\beta}}$. These coefficients are multiplied by $\mathbf{V}$ and passed to the decoder of $M_2$ to reconstruct $\hat{DIC}$.

Training this extended architecture requires approximately five additional minutes compared to the original RRAE. Higher values of $r_{max}$ were tested but showed negligible impact on performance.

\begin{figure}[!h]
\centering
\includegraphics[width=0.4\textwidth]{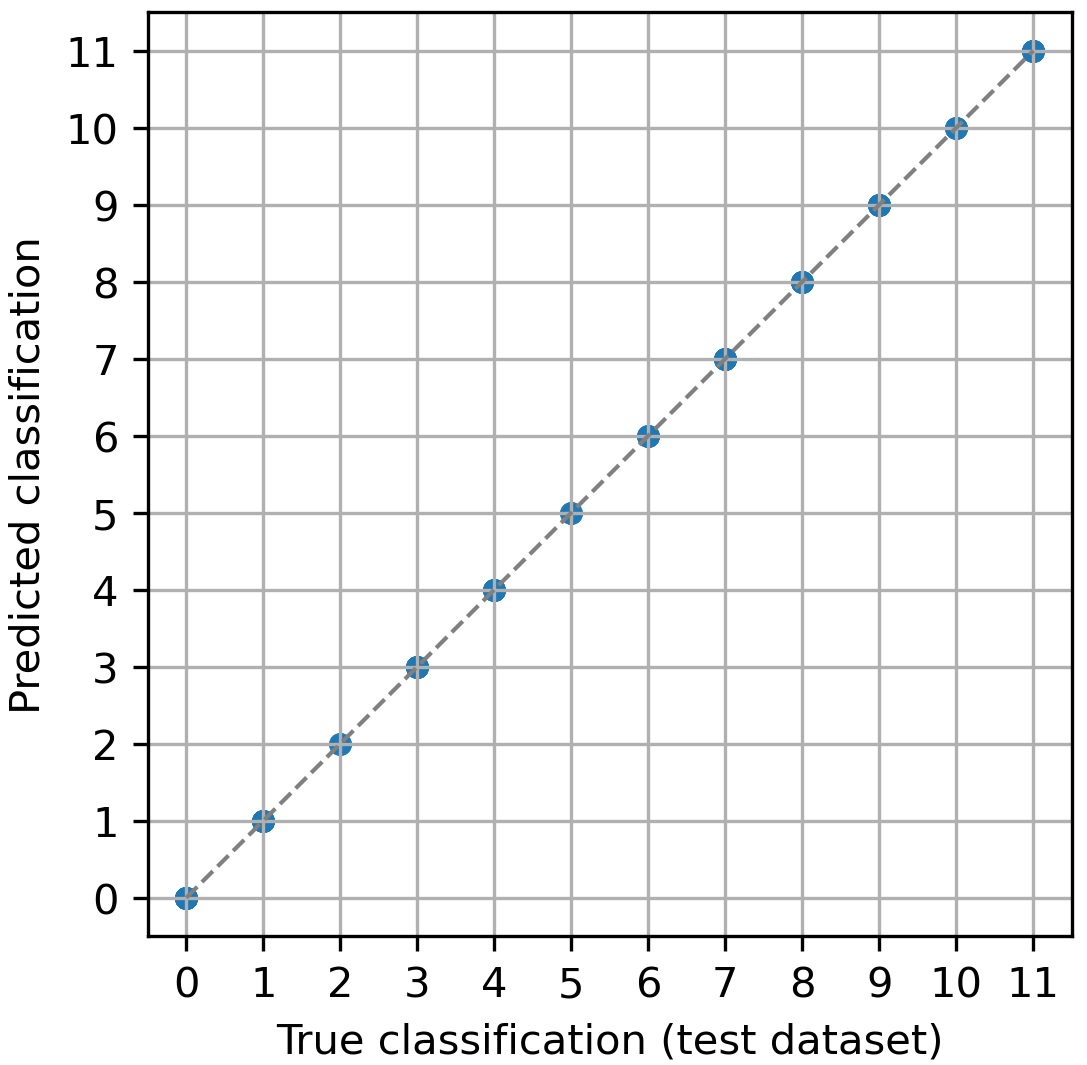}
\caption{Classification results for $k_{max}=4$ using the extended architecture.}
\label{fig:ArchiNew_ClassK4}
\end{figure}

For $k_{max}=4$, the proposed architecture achieves accurate classification (Fig.~\ref{fig:ArchiNew_ClassK4}) while maintaining DIC prediction statistics comparable to the original RRAE (Fig.~\ref{fig:RRAEDicOut_Stat4}). The cumulative DIC error on the test set decreases from 287 (original RRAE with $k_{max}=4$) to 266 with an average error of $2.5\%$. More importantly, classification becomes nearly perfect, in contrast to Fig.~\ref{fig:ClassK4}, where two classes were misidentified.

\begin{figure}[!h]
\centering
        \begin{subfigure}{0.48\textwidth}
            \includegraphics[width=\textwidth]{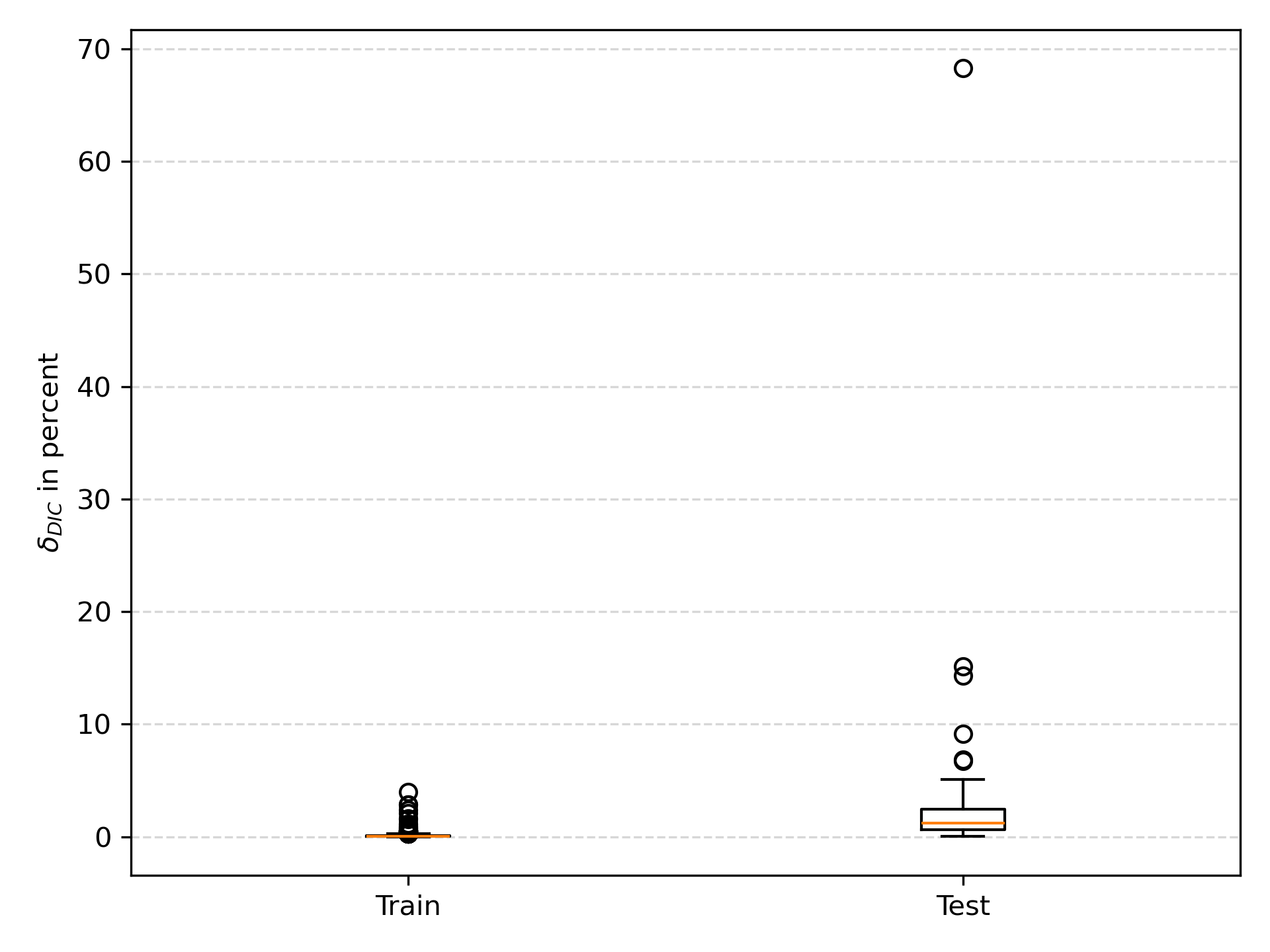}
            \caption{Boxplot of $\delta_{DIC}$}
        \end{subfigure}  
    \hfill
        \begin{subfigure}{0.48\textwidth}
            \includegraphics[width=\textwidth]{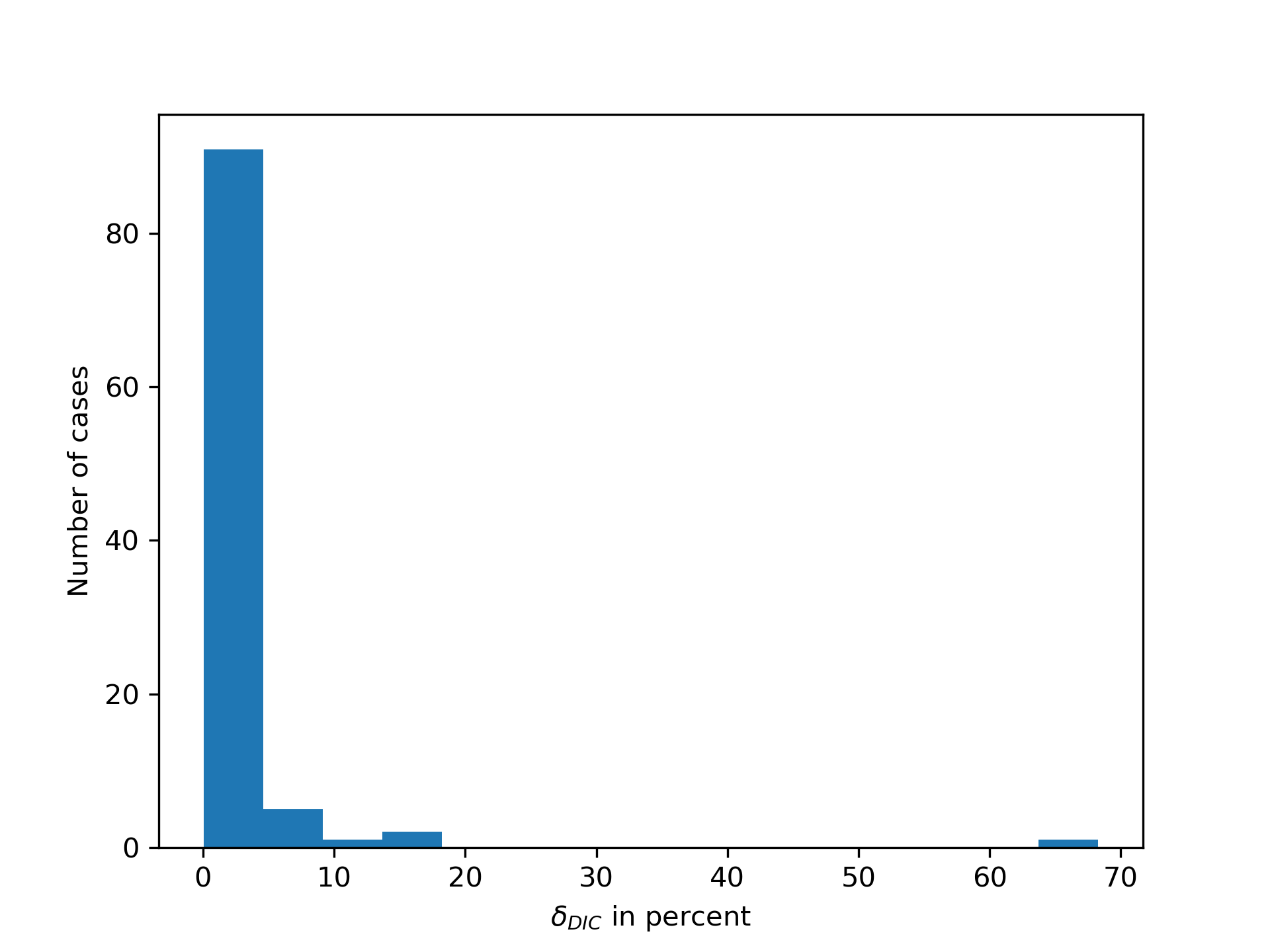}
            \caption{Histogram of test errors}
        \end{subfigure}  
\caption{DIC prediction statistics for the extended architecture with $k_{max}=4$.}
\label{fig:ArchiNew_Stat4}
\end{figure}

\begin{figure}[!h]
\centering
        \begin{subfigure}{0.48\textwidth}
            \includegraphics[width=\textwidth]{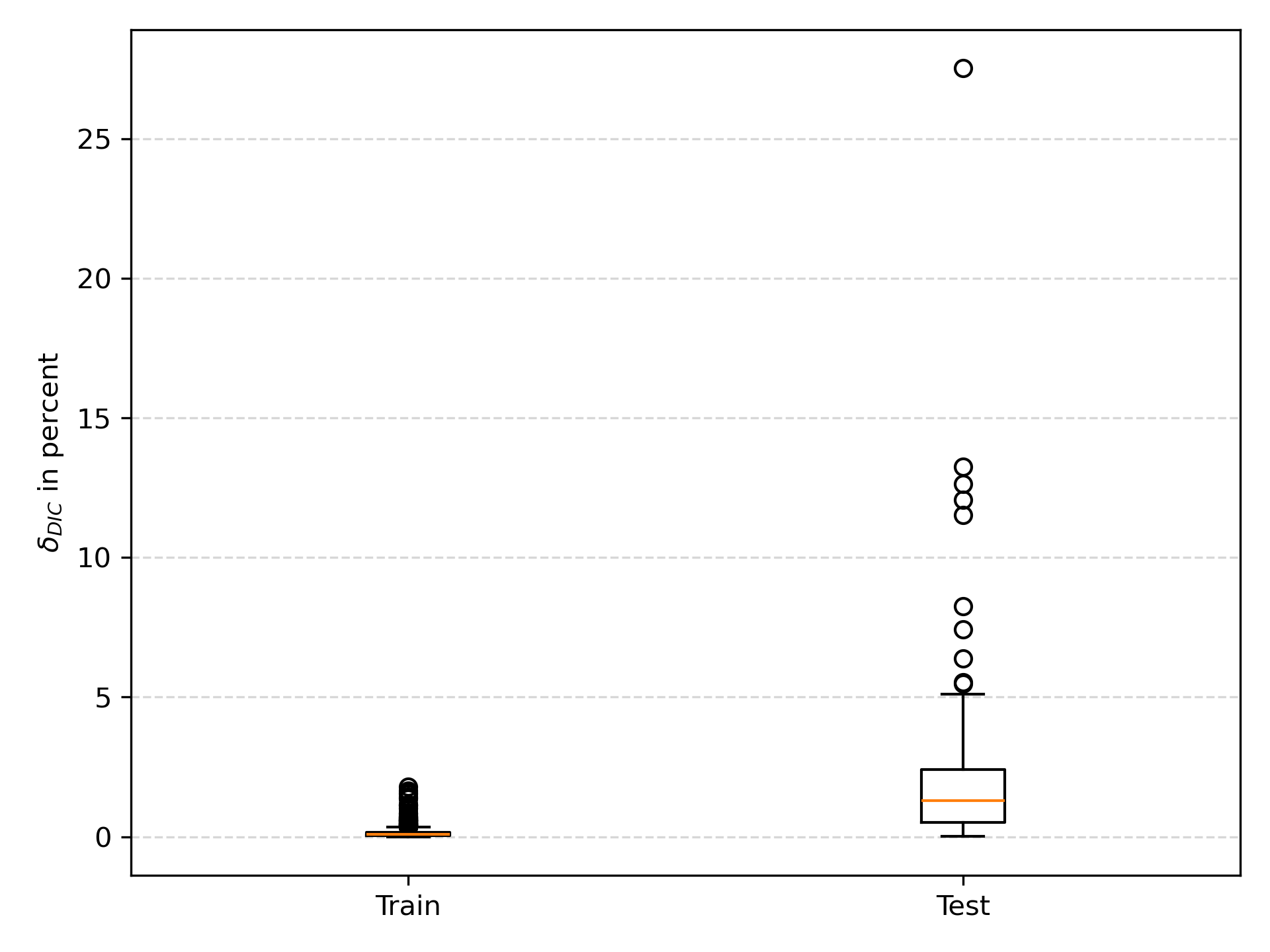}
            \caption{Boxplot of $\delta_{DIC}$}
        \end{subfigure}  
    \hfill
        \begin{subfigure}{0.48\textwidth}
            \includegraphics[width=\textwidth]{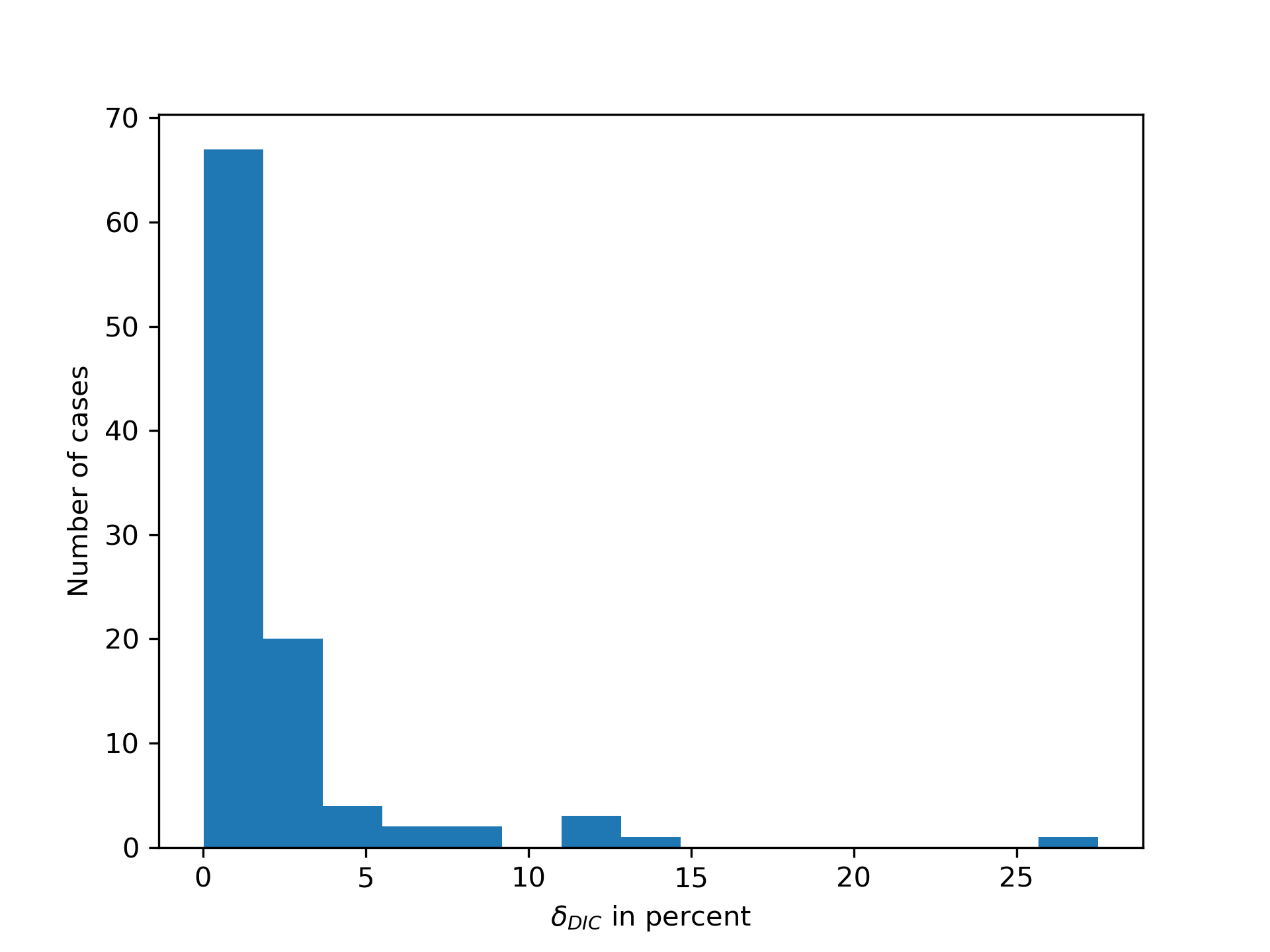}
            \caption{Histogram of test errors}
        \end{subfigure}  
\caption{DIC prediction statistics for the extended architecture with $k_{max}=5$.}
\label{fig:ArchiNew_Stat5}
\end{figure}

Increasing to $k_{max}=5$ further reduces the cumulative DIC error to 220 (Fig.~\ref{fig:ArchiNew_Stat5}), matching the best performance obtained with the original RRAE. Classification performance remains stable. For larger values of $k_{max}$, improvements become marginal.

In summary, this extended architecture demonstrates that four latent roughness features are sufficient to jointly achieve accurate classification and DIC prediction when the intrinsic DIC modes are explicitly separated. Increasing to $k_{max}=5$ provides optimal DIC accuracy. Nevertheless, the additional architectural complexity yields only moderate improvements compared to the original RRAE. Therefore, the architecture of Fig.~\ref{fig:RRAE2MLP} remains the most balanced solution in terms of interpretability, computational cost, and predictive performance.

\section{Comparison with Standard Learning Architectures}\label{sec:MLStand}

One of the main advantages of the RRAE framework lies in its ability to extract relevant latent features from complex roughness profiles. To assess whether this advantage is genuine or merely architectural, alternative state-of-the-art learning strategies are applied to the same task.

Two main challenges characterize the present problem. First, both the input (roughness profile) and the output (DIC evolution) are high-dimensional signals. Although dimensionality reduction techniques can simplify training \cite{tannous2025dthor, yun2025sensor}, such reductions may be incompatible with complex physics and can degrade predictive accuracy. Second, the apparent stochastic nature of the roughness suggests dependence on hidden parameters rather than on controllable input variables suitable for a structured design of experiments.

As a result, fully connected networks typically fail to exploit spatial correlations and tend to overfit or generalize poorly. Standard regression methods such as Random Forest \cite{breiman2001rf}, boosting techniques (XGBoost \cite{breiman1984cart}, CatBoost \cite{prokhorenkova2018catboost}), and Support Vector Machines \cite{cortes1995svm} face similar limitations, since spatial locality is not inherently embedded in their formulation.

Two alternative deep-learning strategies are therefore investigated:

\begin{enumerate}
\item A supervised encoder–decoder architecture mapping the roughness profile directly to the DIC curve.
\item A classical auto-encoder with a purely linear latent compression (without SVD), replacing the modal truncation used in the RRAE.
\end{enumerate}

\subsection{Supervised Encoder--Decoder Network}

Convolutional encoder–decoder architectures explicitly capture translation-invariant local features while organizing information into a structured latent space. Similar approaches, such as UNet \cite{ronneberger2015unet}, have demonstrated strong performance in complex inverse problems \cite{AutoEnc}.  

Here, a compact one-dimensional encoder–decoder is employed to predict the DIC curve from the roughness profile. The encoder consists of two convolutional layers (kernel size 9, stride 3), each followed by max-pooling, progressively reducing spatial resolution while increasing feature maps. The extracted features are flattened and projected onto a latent vector of size $d_\ell$, and then further reduced to a dimension $\gamma$ using a fully connected layer. Several $\gamma$ values where explored, and $\gamma=6$ was retained as further increase in this value contributed to over-fitting.  

The decoder mirrors this structure: latent features are lifted through a fully connected layer, reshaped into feature maps, and upsampled using transposed convolutions to recover the original resolution. A sigmoid activation bounds the output, and dropout regularization is applied throughout. The network contains approximately 6500 trainable parameters.

Increasing model capacity leads rapidly to overfitting, while reducing it significantly degrades prediction accuracy. Architecture tuning therefore becomes a delicate and time-consuming trial-and-error process. In contrast, the RRAE proves less sensitive to encoder–decoder design, with the latent dimension $k_{max}$ being the primary parameter to determine.

\begin{figure}[!h]
\centering
        \begin{subfigure}{0.48\textwidth}
            \includegraphics[width=\textwidth]{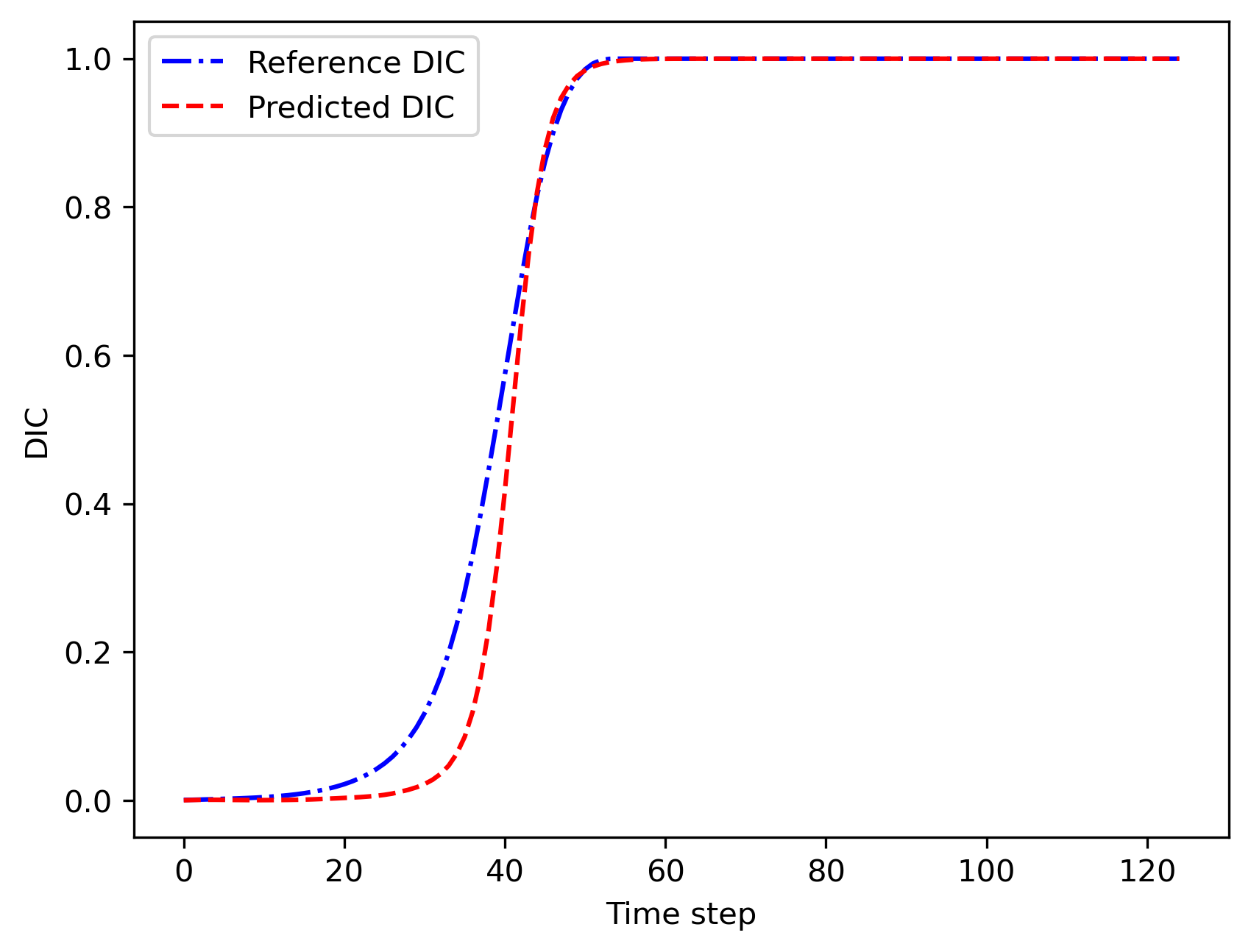}
            \caption{Accurate prediction}
            \label{fig:AutoEnc_DIC1}
        \end{subfigure}  
    \hfill
        \begin{subfigure}{0.48\textwidth}
            \includegraphics[width=\textwidth]{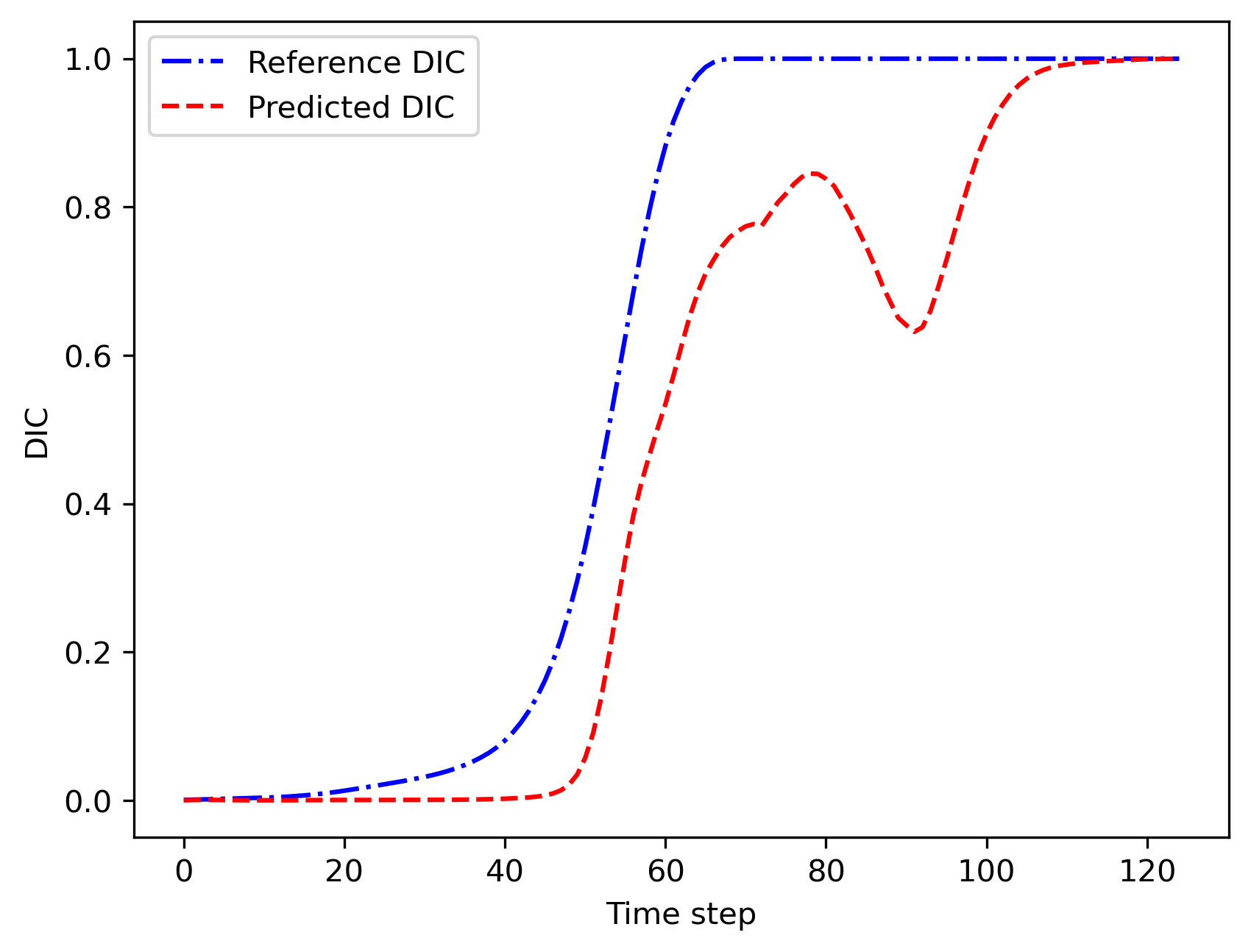}
            \caption{Prediction with spurious oscillations}
            \label{fig:AutoEnc_DIC2}
        \end{subfigure}  
\caption{DIC prediction using a standard encoder–decoder for two representative test roughness profiles.}
\label{fig:AutoEncOut}
\end{figure}

From an accuracy perspective, the supervised encoder–decoder produces globally reasonable DIC predictions. However, owing to the limited size of the dataset, the training process remains highly sensitive to architectural tuning and initialization. Even after careful calibration, spurious numerical oscillations may appear in both training and testing predictions (Fig.~\ref{fig:AutoEnc_DIC2}), leading to non-physical DIC responses. 

When such oscillations are absent, the predictions can be satisfactory (Fig.~\ref{fig:AutoEnc_DIC1}). Nevertheless, even in these favorable cases, the overall accuracy generally remains inferior to that achieved by the RRAE.

\begin{figure}[!h]
\centering
        \begin{subfigure}{0.48\textwidth}
            \includegraphics[width=\textwidth]{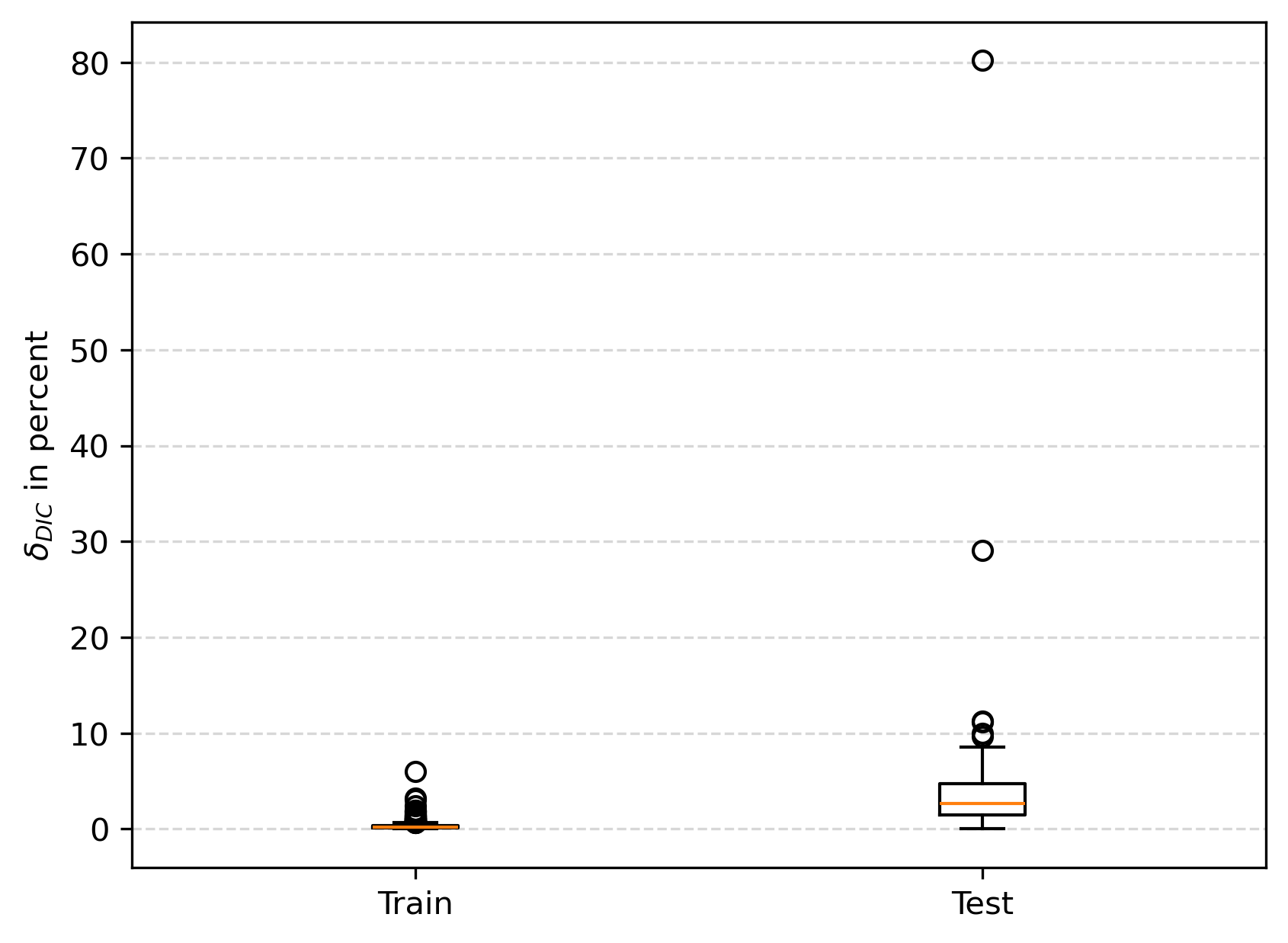}
            \caption{Boxplot distribution of $\delta_{DIC}$}
            \label{fig:Boxplot_EncDCec}
        \end{subfigure}  
    \hfill
        \begin{subfigure}{0.48\textwidth}
            \includegraphics[width=\textwidth]{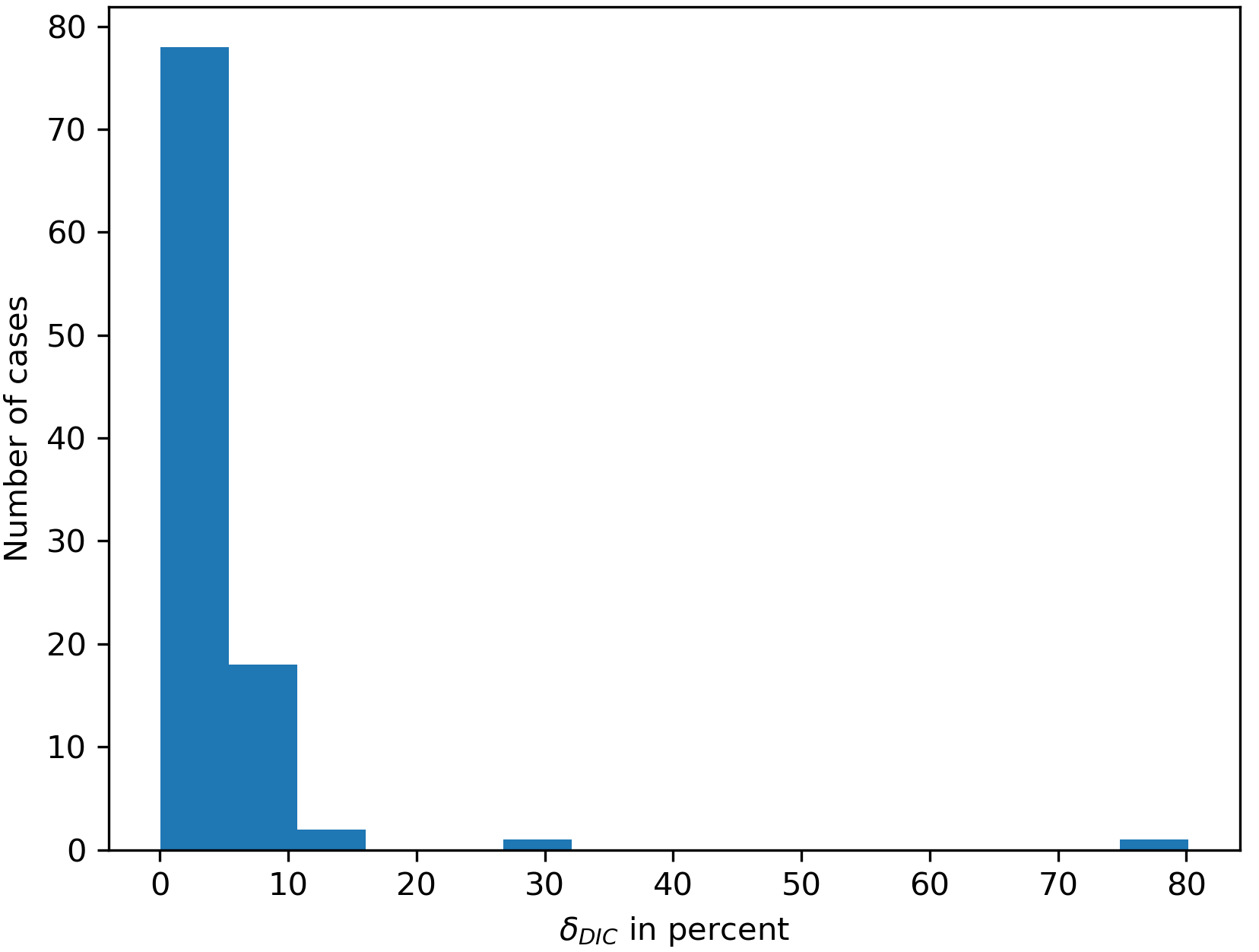}
            \caption{Histogram of test errors}
            \label{fig:HistoTest_EncDec}
        \end{subfigure}  
\caption{Statistical distribution of $\delta_{DIC}$ for the standard encoder–decoder architecture.}
\label{fig:StatEncDec}
\end{figure}

The quantitative error analysis further highlights this difference. Figure~\ref{fig:StatEncDec} presents the distribution of the DIC evolution error $\delta_{DIC}$ for the encoder–decoder architecture. The cumulative $\delta_{DIC}$ over the test set reaches 421, compared to 221 obtained with the RRAE. This corresponds to nearly twice the total error.

It is important to emphasize that this encoder–decoder was trained on a simplified task, focusing solely on DIC prediction and omitting the simultaneous classification objective required in the RRAE framework. Despite this reduced complexity, its predictive performance remains significantly lower. This comparison reinforces the advantage of the RRAE, whose structured latent representation provides improved stability, physical consistency, and overall predictive accuracy.

\subsection{Classical Auto-Encoder with Linear Latent Compression}

A second comparison is carried out using a classical auto-encoder (AE) architecture. The same global structure as the RRAE (Fig.~\ref{fig:RRAE2MLP}) is retained; however, the modal truncation mechanism is replaced by learned linear projections. In this configuration, the encoder reduces the latent dimension from $d_\ell$ to $k_{max}$ through a fully connected layer, followed by a second linear layer that reconstructs the $d_\ell$-dimensional representation. 

All remaining components (activation functions, loss functions, batch size, optimizer, etc.) are kept identical to those used in the RRAE, ensuring a fair comparison. This setup therefore directly questions the necessity and added value of the structured SVD-based reduction employed in the RRAE.

\begin{figure}[!h]
\centering
\includegraphics[width=0.4\textwidth]{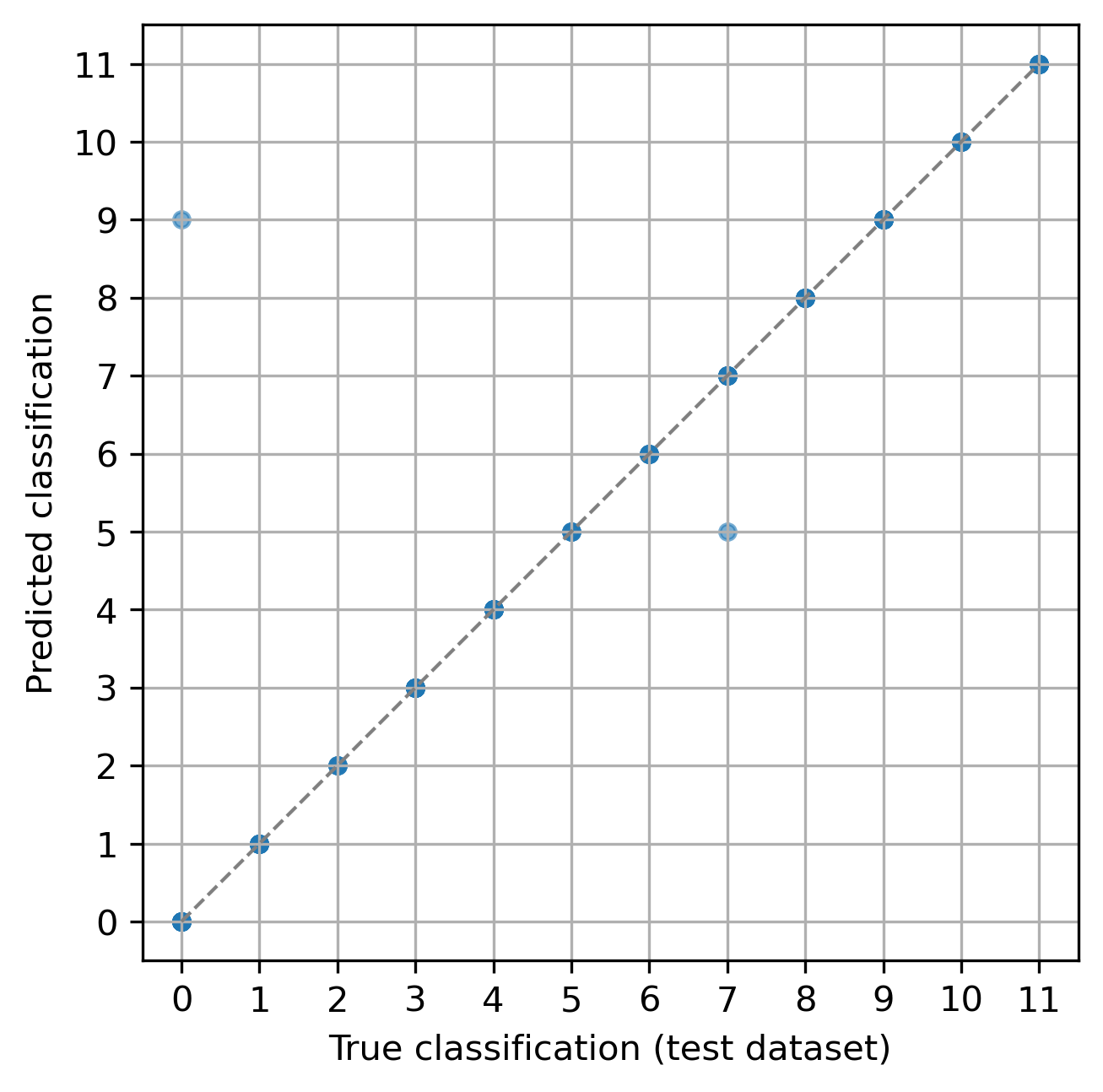}
\caption{Roughness classification using the classical AE architecture for $k_{max}=5$.}
\label{fig:ClassicalAE_kmax5}
\end{figure}

For $k_{max}=5$, the following observations were made:

\begin{enumerate}
\item The classical AE struggled to jointly learn classification and DIC prediction. In particular, classification accuracy was not satisfactory, with two misclassified categories (Fig.~\ref{fig:ClassicalAE_kmax5}).
\item The predicted DIC curves exhibited spurious oscillations, although the global trend was captured reasonably well. The cumulative $\delta_{DIC}$ error on the test set reached approximately 275, which corresponds to roughly $25\%$ higher error than the RRAE (221).
\end{enumerate}

\begin{figure}[!h]
\centering
\includegraphics[width=0.8\textwidth]{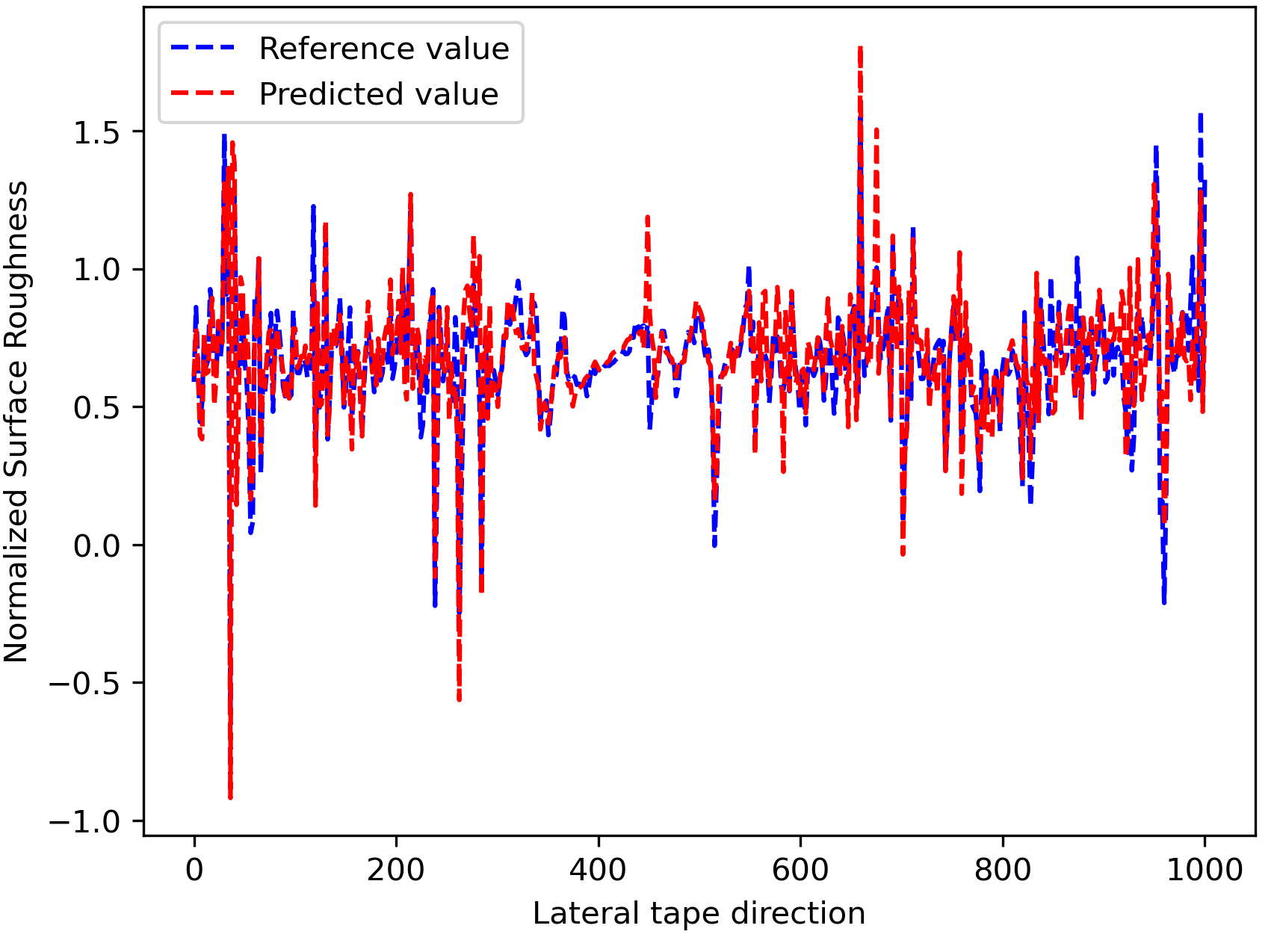}
\caption{Surface roughness reconstruction for a representative test case using the classical AE ($k_{max}=12$).}
\label{fig:StandardAE_Rug}
\end{figure}

\begin{figure}[!h]
\centering
\includegraphics[width=0.4\textwidth]{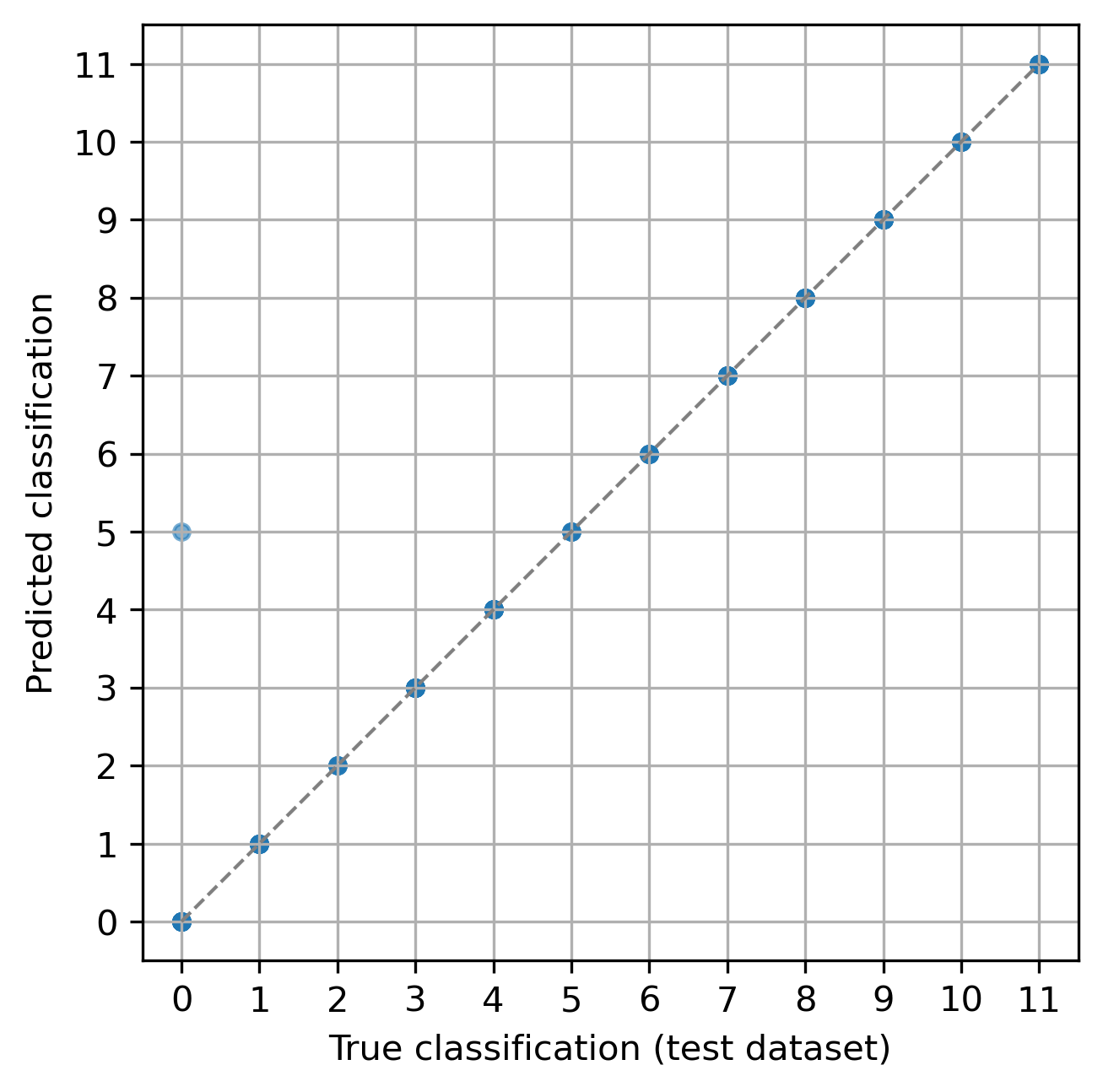}
\caption{Surface roughness classification using the classical AE ($k_{max}=12$).}
\label{fig:Clasification_kmax12}
\end{figure}

To improve performance, $k_{max}$ had to be increased significantly. Satisfactory results were obtained for $k_{max}=12$, beyond which no further improvement was observed. For this value, surface roughness reconstruction was accurate (Fig.~\ref{fig:StandardAE_Rug}). The DIC prediction became comparable to that of the RRAE in terms of global trend and cumulative error. However, classification still exhibited one persistent misclassification (Fig.~\ref{fig:Clasification_kmax12}).

\begin{figure}[!h]
\centering
        \begin{subfigure}{0.48\textwidth}
            \includegraphics[width=\textwidth]{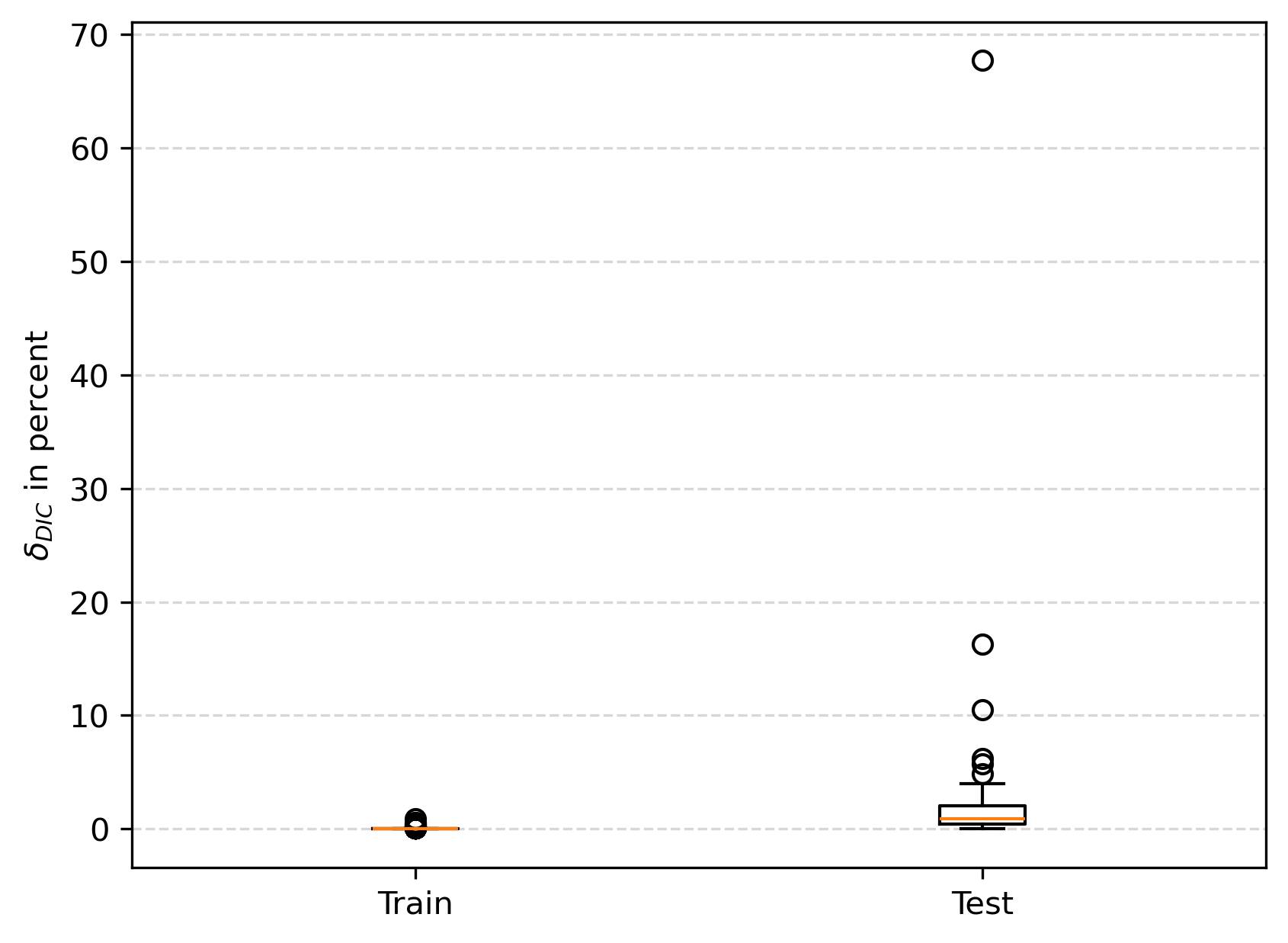}
            \caption{Boxplot distribution of $\delta_{DIC}$}
            \label{fig:AEDicOut_Box6}
        \end{subfigure}  
    \hfill
        \begin{subfigure}{0.48\textwidth}
            \includegraphics[width=\textwidth]{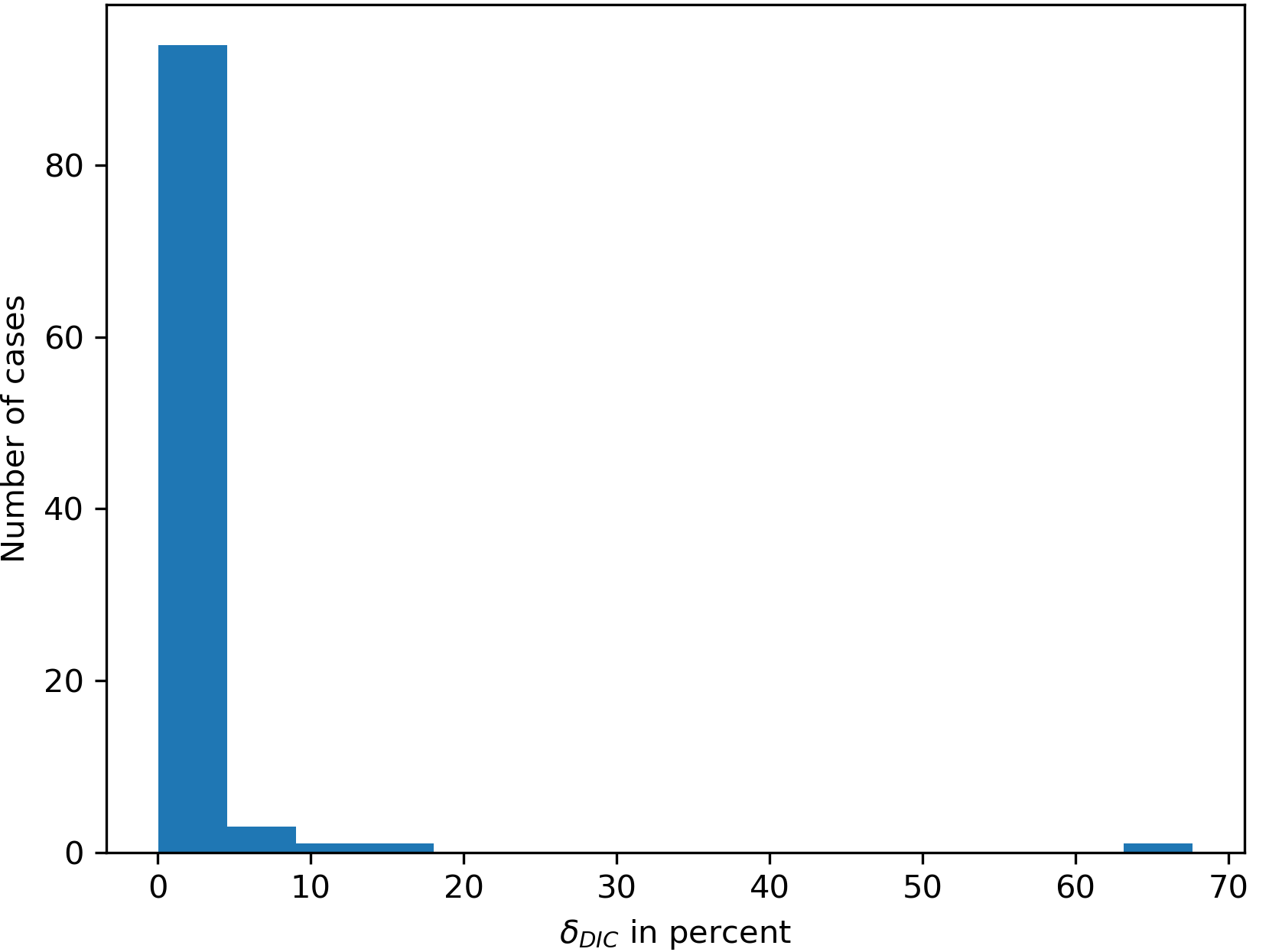}
            \caption{Histogram of test errors}
            \label{fig:AEDicOut_Histo}
        \end{subfigure}  
\caption{Accuracy statistics for DIC prediction using the classical AE ($k_{max}=12$).}
\label{fig:AEDicOut_Stat}
\end{figure}

The statistical distribution of $\delta_{DIC}$ for $k_{max}=12$ is shown in Fig.~\ref{fig:AEDicOut_Stat}. Although the majority of test cases reach error levels comparable to those of the RRAE, the outliers display higher deviations.

Two main conclusions can be drawn from this comparison:

\begin{enumerate}
\item The classical AE can be fine-tuned to approach the performance of the RRAE, but only at the cost of substantially increasing the latent dimension and performing careful architectural adjustments. In contrast, the RRAE achieves higher accuracy with fewer latent features and demonstrates lower sensitivity to architectural tuning. This robustness is particularly advantageous when dealing with limited datasets.
\item The interpretation of $k_{max}$ differs fundamentally between the two approaches. In the RRAE, the enforced SVD structure ensures that the latent features correspond to orthogonal modes, which can be interpreted as representative components of the underlying physics. In the classical AE, the latent space consists of learned linear combinations without modal ordering or orthogonality constraints. Consequently, a larger number of neurons is required, and $k_{max}$ no longer represents a minimal set of physically meaningful features. This difference highlights the interpretability and dimensional efficiency advantages of the RRAE formulation.
\end{enumerate}

\section{Conclusions and Perspectives}\label{sec:conclusion}

This work addressed the analysis of rough composite tape surfaces with two main objectives:
\begin{enumerate}
\item enabling tape classification to support online decision-making and process control based on accumulated experience, experiments, or high-fidelity simulations;
\item estimating compaction ability of composite tapes through real-time prediction of the evolution of the degree of intimate contact (DIC).
\end{enumerate}

To this end, a Rank Reduction AutoEncoder (RRAE) was introduced to extract latent descriptors from the roughness profiles and use them to infer the quantities of interest. Compared with conventional machine learning strategies, the RRAE is easier to design and tune, as the effort focuses primarily on identifying the number of descriptors required to represent the surface. The approach delivered near-perfect classification accuracy and reliable DIC predictions while providing insight into the intrinsic dimensionality of the problem.

Although the physical interpretation of individual descriptors is not explicit, the ability to represent complex roughness using a small and controlled set of variables is a major advantage. Most importantly, the methodology is not tied to the present consolidation model. 
Should the physics be enriched—for instance by describing the press motion through a constant-force rather than constant-velocity formulation—the RRAE framework can be retrained to embed the improved behavior directly into the latent space. In this sense, better simulations or experimental data will immediately translate into more predictive descriptors.

The proposed strategy therefore opens the way toward identifying the hidden drivers of bonding quality, enabling more accurate defect prediction, reduced scrap rates, and enhanced process reliability.

\section*{References}

\end{document}